\documentstyle[12pt,fullpage,psfig]{article}

\newcommand {\gig} {graph interpolation grammar}
\newcommand {\GIG} {Graph Interpolation Grammar}

\newcommand {\GIGs} {Graph Interpolation Grammars}

\newcommand {\paper} {report}

\newcommand {\implies} {~\mbox{$\Rightarrow$}~}
\newcommand {\ancestor} {\mbox{$\Downarrow$}}
\newcommand {\successor} {\mbox{$\Leftarrow$}}
\newcommand {\immediateSuccessor} {\mbox{$\leftarrow$}}

\newcommand {\subsubsubsection}[1] {\paragraph{#1}}

\long\def\startIgnoring #1\stopIgnoring{}
\def\stopIgnoring{}

\newtheorem{definition}{Definition}

\newtheorem{principle}{Principle}
\newtheorem{invariant}{Invariant}
\newtheorem{procedure}{Procedure}

\title{Graph Interpolation Grammars: \\ a rule-based approach
to the incremental parsing of natural languages}
\date{March 1998}
\author{John Larchev\^{e}que}

\startIgnoring
\RRtitle{Les Grammaires \`{a} Interpolations de Graphes\,: \\ un
formalisme d\'{e}claratif pour l'analyse incr\'{e}mentale du discours}

\RRetitle{Graph Interpolation Grammars: \\ a rule-based approach
to the incremental parsing of natural languages}

\titlehead{Graph Interpolation Grammars}

\RRauthor{John Larchev\^{e}que}

\RRdate{March 1998}

\RRtheme{3}
\RRprojet{Atoll}

\URRocq

\RRabstract{Graph Interpolation Grammars are a declarative
formalism with an operational semantics. Their goal is to emulate
salient features of the human parser, and notably
incrementality. The parsing process defined by GIGs incrementally
builds a syntactic representation of a sentence as each
successive lexeme is read. A GIG rule specifies a set of parse
configurations that trigger its application and an operation to
perform on a matching configuration. Rules are partly
context-sensitive; furthermore, they are reversible, meaning that
their operations can be undone, which allows the parsing process
to be nondeterministic. These two~factors confer enough
expressive power to the formalism for parsing natural languages.}

\RRkeyword{parsing, natural language
processing, psycholinguistics, linguistics, grammar, syntactic
representation, lexical representation}

\RRresume{Les Grammaires \`{a} Interpolations de Graphes sont un
formalisme d\'{e}claratif muni d'une s\'{e}mantique
op\'{e}rationnelle. Elles visent \`{a} \'{e}muler certaines
caract\'{e}ristiques de l'analyseur syntaxique humain, et
notamment son caract\`{e}re incr\'{e}mental. Le processus
d'analyse qu'elles d\'{e}finissent construit une
repr\'{e}sentation syntaxique incr\'{e}mentalement, \`{a} mesure
que chaque lex\`{e}me est lu.  Dans une GIG, une r\`{e}gle
comporte une repr\'{e}sentation des configurations
d\'{e}clenchantes et une sp\'{e}cification des op\'{e}rations
\`{a} appliquer sur ces configurations. Les r\`{e}gles comportent
un \'{e}l\'{e}ment contextuel; de plus, elles sont
r\'{e}versibles, en ce sens qu'il est possible d'en d\'{e}faire
les effets, ce qui permet une analyse non d\'{e}terministe. Ces
deux facteurs contribuent \`{a} conf\'{e}rer aux Grammaires \`{a}
Interpolations de Graphes un pouvoir d'expression suffisant pour
l'analyse des langues naturelles.}

\RRmotcle{syntaxe, langage
naturel, psycholinguistique, linguistique, grammaire, repr\'{e}sentation
syntaxique, repr\'{e}sentation lexicale}
\stopIgnoring

\begin{document}


\maketitle

\abstract{Graph Interpolation Grammars are a declarative
formalism with an operational semantics. Their goal is to emulate
salient features of the human parser, and notably
incrementality. The parsing process defined by GIGs incrementally
builds a syntactic representation of a sentence as each
successive lexeme is read. A GIG rule specifies a set of parse
configurations that trigger its application and an operation to
perform on a matching configuration. Rules are partly
context-sensitive; furthermore, they are reversible, meaning that
their operations can be undone, which allows the parsing process
to be nondeterministic. These two~factors confer enough
expressive power to the formalism for parsing natural languages.}
\section{Introduction}

\subsection{Characteristics and rationale}

A \gig\ is a grammar formalism with an operational semantics. A
rule in a \gig\ specifies not only syntactic relations but also
an elementary parsing operation.

Rules are lexicalized in the sense that each rule describes the
combinatory properties of a lexical item. Parsing a sentence
consists in matching each lexeme in the input string with a rule
in the grammar and applying this rule to the current parse
representation.

GIG-driven parsing was designed with incrementality and
flexibility in mind, so as to emulate some features of the human
parsing capability, in particular incremental processing, error
tolerance, and handling of complex word orders.

In a complete model of discourse understanding, incremental
parsing would be shown to work in tandem with some form of
composition between partial semantic representations. It is to be
thought that the collaboration between syntax and semantics in
natural discourse understanding is fairly close, and that
backtracking in the parser is frequently initiated by a clash
found between semantic features. This \paper, however, will not
attempt to give even the roughest idea of what semantic
representations should look like and will therefore exclusively
focus on syntactic phenomena, sometimes at the cost of
simplifying the phenomena at hand.

\subsection{Plan of the \paper}

The first two sections give a fairly complete presentation of the
concepts and processes involved in parsing with a \GIG.  The
syntactic structures generated when parsing with a GIG are
described in Section~\ref{syntacticStructures.sec}, while the
grammar rules and the parsing process are described in
Section~\ref{buildingParseGraphs.sec}.

The next few sections apply the formalism to selected problems.
Section~\ref{cfg.sec} compares the handling of a simple
expression language using a Context Free Grammar and a
\GIG, and Section~\ref{DCSD.sec}
analyzes Dutch Cross-Serial Dependencies.

Finally, the conclusion indicates topics for further research and
related works.

\section{Syntactic structures}
\label{syntacticStructures.sec}

\subsection{Phrases}

A phrase is made up of a head and its complements. It will be
represented by a graph with a root and one labelled edge from the
root to each nonroot node. 

The root represents the head of the phrase, the other nodes
represent its complements; and the edge labels represent
grammatical functions.

\subsubsection{Ordering relations}

In addition, nodes in a phrase graph are connected through {\em
ordering edges}. In the simplest case, the effect of ordering
edges is to align graph nodes from left to right.

For example, the sentence 
\begin{quote}
She gave me an apple.
\end{quote}
contains a phrase the head of which is a ditransitive verb with
three complements, the subject, the indirect object, and the direct
object. This phrase can be represented by the following graph (in
which {\em V} for {\em verb} and {\em NP} for {\em noun phrase}).

\begin{figure}[htbp]
\begin{center} ~\psfig{file=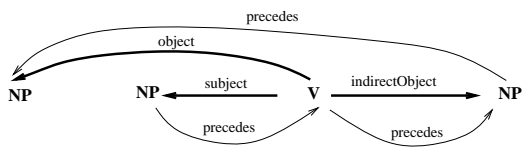,height=2cm} \end{center}
\caption{A phrase graph}
\label{phrase1.fig}
\end{figure}

\begin{figure}[htbp]
\begin{center} ~\psfig{file=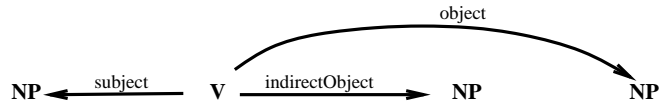,height=15mm} \end{center}
\caption{The same phrase graph with implicit ordering}
\label{phrase2.fig}
\end{figure}

Since this is a case in which a total left-to-right alignment of
nodes can be defined, ordering edges can be omitted and the
ordering represented by the relative positions of the nodes in
the diagram, as in Figure~\ref{phrase2.fig}. Explicit orderings
will not be used in this \paper, for a strict total ordering of
phrase constituents exists in all examples given. At any rate,
the possibility exists of specifying sophisticated order
constraints as long as these constraints are local to a phrase.

\stopIgnoring

\subsubsection{Phrase head}

The criteria for distinguishing a head in a phrase are both
syntactic and semantic. From a syntactic point of view, the head
determines the presence and functions of its complements. From a
semantic point of view, the head acts as a predicate of its
complements, adding semantic features to them, while the
complements generally contribute to rooting these features into a
specific situation.

\begin{figure}[htbp]
\begin{center} ~\psfig{file=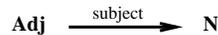,height=16pt} \end{center}
\caption{A modifying phrase}
\label{phrase3.fig}
\end{figure}

On the basis of these criteria, an attributive adjective forms
the head of a phrase made up of an adjective and a noun, such as
{\em red leaf}. Indeed, the occurrence of the adjective implies
the occurrence of a single noun, whereas
the presence of the noun does not require or limit the number of
attributive adjectives. On the other hand, though evidence for
this type of criterion is not so easy to adduce, the adjective
adds semantic features to the noun, as is shown in particular by
its capacity to function as a predicate in a
clause. Figure~\ref{phrase3.fig} illustrates a phrase headed by
an attributive adjective. ({\em Adj} stands for {\em adjective}.)

Considering the adjective as the head of such a phrase can be
counterintuitive insofar as the phrase in question functions as a
noun. It is, however, necessary to distinguish the internal
structure of a phrase ---as described by its phrase graph--- from
its function with respect to other phrases. In order to bring out
this distinction, a structure that interrelates phrase graphs
must be defined. The next section defines a structure of this
type, the {\em parse graph}.

\subsection{Parse graphs}
\label{parseGraphs.sec}

A parse graph is a structure in which phrase graphs are related
through {\em parent-of} edges.

\begin{figure}[htbp]
\begin{center} ~\psfig{file=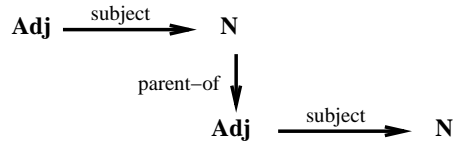,height=2cm} \end{center}
\caption{A parse graph}
\label{phrase4.fig}
\end{figure}

Consider for example the phrase {\em good old days}.  It
contains the phrase {\em old days}, which has {\em old} as its
head, but functions as a noun and is itself modified by the
adjective {\em good} in the larger phrase {\em good old
days}. The 'containment' relation just outlined can be rendered
by drawing a {\em parent-of} edge between the noun node of the
enclosing phrase and the head of the embedded
phrase. Graphically, this can be represented as in
Figure~\ref{phrase4.fig}.

From now on, in parse graph representations, vertical edges will
be assumed to be directed downward and represent {\em parent-of}
edges.

On the other hand, a {\em parent-of} edge can also connect a
phrase node to a lexical item. A complete parse graph for a
phrase or sentence can thus be defined as a parse graph with a
root node and such that every path leads to a lexical item. These
facts give rise to the following principle and definition.

\begin{principle}
\label{lexicalNode.princ}
A lexical node can occur in a parse graph as the destination of a
{\em parent-of} edge.
\end{principle}

\begin{definition}
\label{completeParseGraph.def}
A {\em complete parse graph} is a parse graph with a root node,
i.e. a node from which every node is reachable, and such that
every path leads to a lexical node.
\end{definition}

\begin{figure}[htbp]
\begin{center} ~\psfig{file=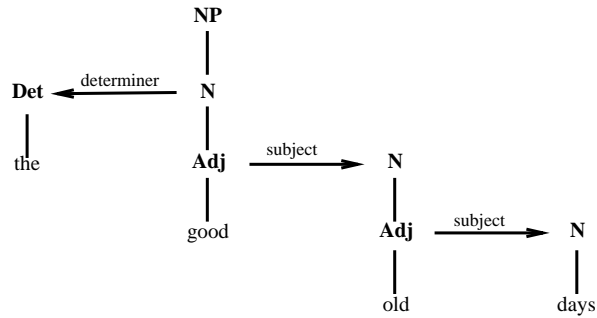,height=42mm} \end{center}
\caption{A complete parse graph}
\label{phrase5.fig}
\end{figure}

Figure~\ref{phrase5.fig} shows the complete parse
graph for the phrase {\em the good old days}. ({\em Det}
stands for {\em determiner}.)

Such a graph can be mapped to a tree in such a way that members
of a phrase graph, including the head, are mapped to sibling
nodes. Although the graph contains more information and is better
adapted to incremental construction that its tree counterpart, it
will often prove useful to apply tree terminology to parse
graphs. For example, the terms {\em descendant}, {\em ancestor},
or {\em frontier} are to be understood with respect to the tree
counterpart of a parse graph. Likewise, a {\em dangling node} is
to be understood as a non-lexical node that occupies a terminal
position in the tree counterpart of an incomplete parse graph.
For further reference, here is the definition that justifies this
terminology. 

\begin{definition}
\label{counterpartTree.def}
The {\em summary tree} of a parse graph is a connected graph
that contains the same set of nodes as the parse graph and such
that each path of the parse graph consisting of either a {\em
parent-of} edge or a {\em parent-of} edge followed by a
functional edge is mapped to a {\em parent-of} edge of the
summary tree.
\end{definition}

\begin{figure}[htbp]
\begin{center} ~\psfig{file=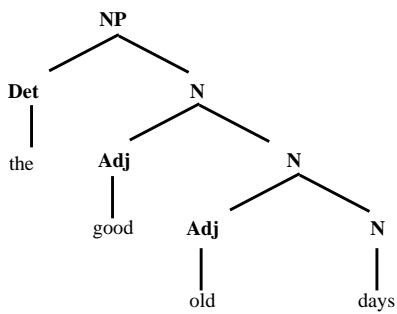,height=42mm} \end{center}
\caption{Summary tree for the parse graph on Figure~\protect\ref{phrase5.fig}}
\label{phrase6.fig}
\end{figure}

\subsection{Subtyping in node labels}
\label{subsumption.sec}

The classification of lexical items can be made according to
several possible criteria. For example, {\em the} can be
categorized as an article and thus be distinguished from {\em
this}. On the other hand, both of {\em the} and {\em this} are
definite determiners and, as such, can be usefully grouped under one
category. This suggests that lexical categories are usefully
viewed as sets of syntactic properties, such as ``determiner'',
``article'', ``definite'', etc. Furthermore, some of these
properties are visible in ancestors of lexical items in a parse
graph. For example, the presence of a definite determiner confers
to a Noun Phrase the status of a definite description. A simple
way of modeling this consists in labelling each node of a parse
graph with a set of property names rather than an atomic
name. Thus, the label of a node dominating {\em the} is a set
containing the names ``determiner'', ``article'', and
``definite'', and the label of an NP determined by {\em the},
{\em this}, a genitive NP, or any other definite determiner, is a
set containing the name ``definite''.

For simplicity, most of the graphs and rules given as examples
have atomic node labels, even when some node types could be shown
to subsume or share properties with other symbols. However,
Section~\ref{expressionWithInheritance.sec} shows how
multi-valued node labels can contribute to the expressive power
of a grammar.

\section{Building parse graphs}
\label{buildingParseGraphs.sec}

Parse graphs, as just defined, can be incrementally built by
scanning an input flow of lexemes and performing a building step
as each lexeme is read. A building step is formally described by a
{\em composition rule}, which adds to the graph under construction a
subgraph associated with the current lexeme.

\subsection{Form of a composition rule}
\label{rule.sec}

\begin{figure}[htbp]
\begin{center} ~\psfig{file=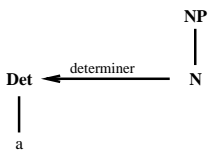,height=2cm} \end{center}
\caption{An NP context}
\label{npContext.fig}
\end{figure}

Suppose that {\em (i)}~the phrase being parsed is {\em a robin},
{\em (ii)}~somehow (see Section~\ref{initialization.sec}), the
parse graph on Figure~\ref{npContext.fig} was built on
encountering the determiner~{\em a}, and {\em (iii)}~the current
lexeme is {\em robin}.

The presence of the dangling ---i.e. childless--- noun node to
the right of the previous lexeme, namely {\em a},
materializes the anticipation of a noun. Singular common nouns
are generally inserted in just such a context, and so we can
define a rule for the lexeme {\em robin} that stipulates {\em
(i)}~that the linear successor of the previous lexeme (i.e. {\em a})
in the parse graph should be a dangling noun node, {\em (ii)}~that the
new subgraph to integrate consists of a noun node dominating the
lexeme {\em robin}, and {\em (iii)}~that this new subgraph 
is to be substituted for the dangling noun node. (A
graphic representation of this rule is given on
Figure~\ref{graphicRule1.fig}, in
section~\ref{insertionExample.sec}.)

More generally, let the {\em context} be the parse graph obtained
by parsing the input string up to the current lexeme, a
composition rule consists of:
\begin{enumerate}
\item a pattern to match against the context, known as the {\em
context pattern}, and
\item a subgraph to integrate into the context, known as the {\em
addendum}. 
\end{enumerate}

Or, more formally:

\begin{definition}
\label{compositionRule.def}
A composition rule is a pair made up of a context pattern and an
addendum. 
\end{definition}

This definition in turn relies on the definitions of a context
pattern and an addendum.  The following definition of a context
pattern is somewhat simplified and will be enriched later
(Definition~\ref{contextPattern.def}, Section~\ref{KleeneStar.sec}).

\begin{definition}
\label{simpleContextPattern.def}
A {\em context pattern} is a parse graph with one distinguished
node called the {\em context anchor} and an indication of whether
the context anchor is an ancestor or immediate linear successor
of the previous lexeme. 
\end{definition}

To proceed in depth-first order, here is the definition that
explains the term {\em linear successor}.

\begin{definition}
\label{linearSuccessor.def}
When all phrase constituents in a parse graph are strictly
ordered from left to right, a {\em linear order} over the parse
graph is given by the left-to-right ordering of nodes in the
frontier of the summary tree
(Definition~\ref{counterpartTree.def},
Section~\ref{parseGraphs.sec}). It is therefore a partial
ordering of parse graph nodes, under which only terminal nodes of
the summary tree are comparable.
\end{definition} 

The context pattern indicates what the parse graph obtained so
far must look like in order for the rule to be applicable. A
context matches a context pattern if it is a supergraph of the
pattern and it contains a node whose location and label match the
anchor specifications.

Section~\ref{freeWordOrder.sec}, on free word order, examines the
consequences of having several possible linearizations of the
parse graph. This has computational implications but does not
impact the above definitions in any essential way.

\begin{definition}
\label{addendum.def}
An {\em addendum} is a parse graph with one or two distinguished
nodes called {\em addendum anchors} and exactly one lexical node.
\end{definition}

A rule is found applicable when the current lexeme matches the
addendum lexeme and the context matches the context pattern. The
conjunction of context and current lexeme forms a {\em parse
configuration}. 

\begin{definition}
\label{configuration.def}
A {\em parse configuration} is a pair made up of a context and a
current lexeme.
\end{definition}

Beside a parse configuration, what a rule specifies is an
operation to modify this configuration.  This operation, called
an {\em interpolation}, is specified by means of the context
anchor and the addendum.

When the addendum contains a single anchor, such as the noun node
in our example, the interpolation is performed by substituting
the addendum anchor for the context anchor. This particular type
of interpolation will sometimes be called an {\em insertion}, by
opposition to a {\em proper interpolation}, which is
characterized by the presence of two~distinct anchors in the
addendum.

When the addendum contains two anchors, they delimit a path that
is substituted for the context anchor. Since a single anchor can
be viewed as a path of length~0, this case subsumes insertion.

No formal definition of interpolation will be attempted at
present. Let it simply be clear that interpolation involves in
fact two operations:
\begin{enumerate}
\item the substitution of a path of the addendum for the context
anchor, 
\item the disjoint union of the addendum with the context.
\end{enumerate}

These notions will be treated in more detail in
Section~\ref{substitutionRevisited.sec}.

\subsection{Insertion example}
\label{insertionExample.sec}

\begin{figure}[htbp]
\begin{center} ~\psfig{file=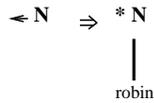,height=15mm} \end{center}
\caption{An insertion rule}
\label{graphicRule1.fig}
\end{figure}

An example of insertion is the example given informally above.
The rule can be represented as on Figure~\ref{graphicRule1.fig}.

In a graphic representation of a composition rule, the context
pattern and the addendum appear on either side of an implies
($\implies$)~sign, with the addendum on the right; and the
anchoring specifications are materialized by placing a mark next
to each anchor. This mark is an asterisk~($*$) next to an
addendum anchor. Next to the context anchor, it is a double down
arrow (\ancestor) or a left arrow (\immediateSuccessor) according
as the context anchor is an ancestor or an immediate successor of
the previous lexeme. The notion of {\em immediate successor} is
defined with respect to the linear order indicated through the
ordering edges present in addenda. Its applicability to free
word-order languages is discussed in
Section~\ref{freeWordOrder.sec}.

\begin{figure}[htbp]
\begin{center} ~\psfig{file=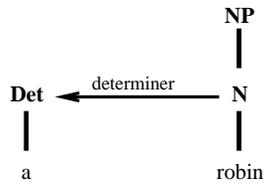,height=25mm} \end{center}
\caption{Result of applying the above insertion rule}
\label{completeNp.fig}
\end{figure}

Applying the insertion rule on Figure~\ref{graphicRule1.fig} to
the parse graph on Figure~\ref{npContext.fig} results in the
parse graph on Figure~\ref{completeNp.fig}.

\subsection{Proper interpolation example}

\begin{figure}[htbp]
\begin{center} ~\psfig{file=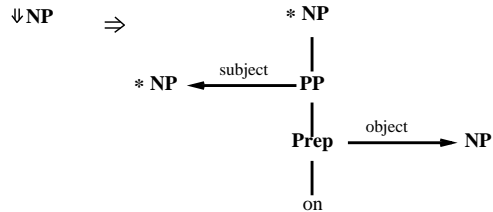,height=30mm} \end{center}
\caption{A proper interpolation rule}
\label{graphicRule2.fig}
\end{figure}

Figure~\ref{graphicRule2.fig} illustrates path-to-node anchoring.
The composition rule interpolates a prepositional
phrase into a noun phrase to produce a qualified noun phrase as in
{\em a bird on a tree}. ({\em PP} stands for {\em Prepositional
Phrase} and {\em Prep} for {\em preposition}.)

\begin{figure}[htbp]
\begin{center} ~\psfig{file=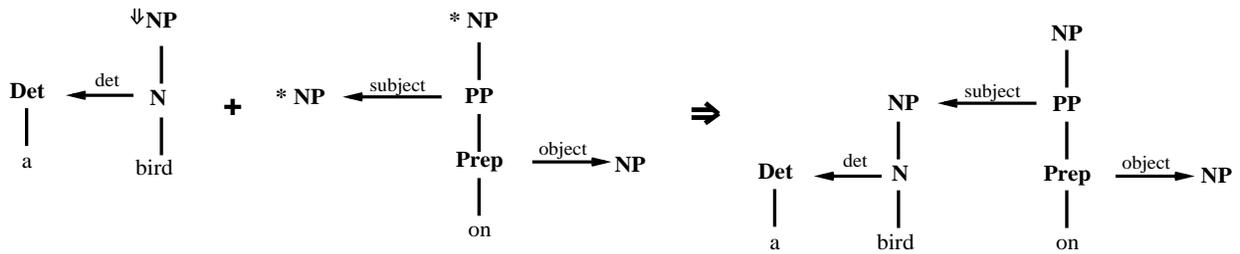,width=\textwidth} \end{center}
\caption{An application of the preceding rule}
\label{applic1.fig}
\end{figure}

The sign next to the context anchor indicates that it is to
be found above the previous lexeme rather than immediately
after it.  The diagram on Figure~\ref{applic1.fig} shows the
context to which this rule could apply and the resulting context
after its application.

An application diagram follows the following pattern.
\begin{displaymath}
\mbox{context~} + \mbox{~addendum~} \implies \mbox{~new~context}
\end{displaymath}

\subsection{Anchor matching}
\label{basicAnchorMatching.sec}

The preceding examples have shown that the context anchor can be
an ancestor or an immediate linear successor of the previous
lexeme. The first case occurs typically with a proper
interpolation, and the second with an insertion. This is, of
course, just the typical case; Section~\ref{anchorMatching.sec}
gives general criteria for locating the context anchor.

\subsubsection{Matching an immediate successor anchor}

No attempt to deal with free word order will be made until
Section~\ref{anchorMatching.sec}. Until then, we can rely on a
strict left-to-right ordering of all phrase constituents.
Therefore, the previous lexeme always has a unique immediate
linear successor, which is its immediate successor in the
frontier of the summary tree
(Definition~\ref{counterpartTree.def},
Section~\ref{parseGraphs.sec}).

\subsubsection{Matching an ancestor anchor}

Ancestor positions are also defined with respect to
the summary tree of the parse graph 
(Definition~\ref{counterpartTree.def},
Section~\ref{parseGraphs.sec}). On the other hand, for a node to
match an ancestor anchor there is an additional condition beside
location and label match. It can be stated as follows.

\begin{principle}
\label{ancestorMatching.princ}
A context anchor in ancestor position can only be matched by the
root of a complete parse graph (see
Definition~\ref{completeParseGraph.def},
Section~\ref{parseGraphs.sec}).
\end{principle}

For example, an NP cannot be postmodified if it is incomplete,
i.e. if the frontier of its summary tree contains nonlexical
nodes. 

\subsubsection{Incidence of node subtyping on anchor matching}

If node labels are sets of symbols, label matching requires that
all symbols attached to the context anchor be present in the node
to match. For example, a context pattern could contain the
information that the anchor is a definite determiner, and
any node that has at least the type atoms ``definite'' and
``determiner'' will satisfy label matching.

In other words, for a node to match a context anchor, its type
must be a subtype of the type of the context anchor. On the other
hand, this means that the formulation of a rule, and notably the
addendum, must involve a special device to mention the full type
of the node that matched the anchor, so as not to lose
information. (Indeed, a rule is allowed to change the type of the
anchor, as explained in Section~\ref{initialization.sec} and
illustrated on Figure~\ref{flewInterpolation.fig}.)
Accordingly, the special symbol {\em \&} can be used
to assign to an addendum anchor the exact type of the node
matching the context anchor.

An example of a grammar that uses subtyping is given in
Section~\ref{expressionWithInheritance.sec}.

\subsection{Parser initialization}
\label{initialization.sec}

To begin with, there is no context or, rather, the context
consists of a single node with a generic label that subsumes all
phrases with which an utterance is likely to start\footnote
{Label subsumption is discussed in
Section~\ref{subsumption.sec}.}. 
No lexeme has been integrated to the context yet; but, for
technical purposes, the initial node can be considered to ``follow
the previous lexeme''. In other words, a rule with an
immediate-successor~(\immediateSuccessor) mark next to the
context anchor will match the initial context.

Suppose a sentence begins with the phrase {\em a robin}; then,
the initial lexeme, {\em a}, can be integrated using the
following insertion rule.

\begin{figure}[htbp]
\begin{center} ~\psfig{file=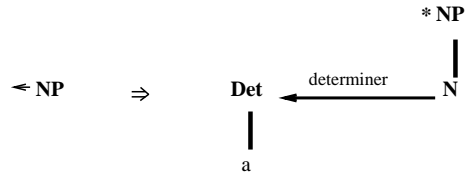,height=25mm} \end{center}
\caption{Insertion rule for {\em a}}
\label{insertA.fig}
\end{figure}

\begin{figure}[htbp]
\begin{center} ~\psfig{file=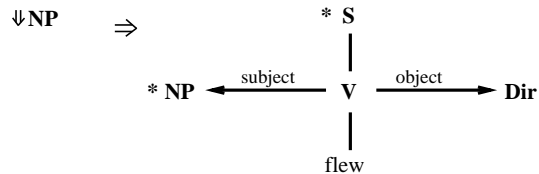,height=25mm} \end{center}
\caption{Interpolation rule for {\em flew}}
\label{flewInterpolation.fig}
\end{figure}

This rule simply means that {\em a} can be considered as the
beginning of a Noun Phrase. It does not say that {\em a} could be
the beginning of a sentence. Yet, most utterances are sentences;
so it would seem that this rule is generally inappropriate. It is
not, however, if one considers that the definition of an
interpolation that was (informally) given in
Section~\ref{rule.sec} does not require the two ends of an
interpolation path to bear the same label. And, indeed, should a
phrase like {\em a robin} be followed with a verb, such as {\em
flew}, the rule shown on Figure~\ref{flewInterpolation.fig} can
apply to make the root of the parse graph an {\em S}-node. ({\em
Dir} stands for ``directional phrase''.)

\subsection{Backtracking}
\label{backtracking.sec}

Matching a rule often involves making a prediction on the continuation
of the utterance. For example, the rule on
Figure~\ref{flewInterpolation.fig} predicts that a directional
phrase follows the verb {\em flew}.
When several predictions are possible, as will
often be the case, a trial-and-error approach is adopted.

A prediction conflict appears when several rules are available
for a given context and current lexeme (i.e. a given parse
configuration). It is supposed that the rules associated with
each lexeme are totally ordered by priority, and each rule is
tried in priority order until a successful parse is
obtained\footnote
{If the selection process parallels a similar process in the
human mind, a simple hypothesis would correlate rule priority
with statistical efficiency; alternately, or complementarily,
there could be structural criteria that correlated priority with
simplicity or canonicity.}.

For example, when the current lexeme is {\em flew} and an NP~node
dominates the previous
lexeme, there could be two rules,
with higher priority on the rule on
Figure~\ref{flewInterpolation.fig}, and lower priority on the
following rule, which makes {\em flew} an intransitive verb.

\begin{figure}[htbp]
\begin{center} ~\psfig{file=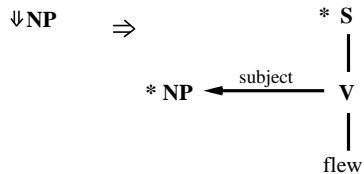,height=25mm} \end{center}
\caption{An alternative rule for {\em flew}}
\label{alternativeFlewInterpolation.fig}
\end{figure}

\begin{figure}[htbp]
\begin{center} ~\psfig{file=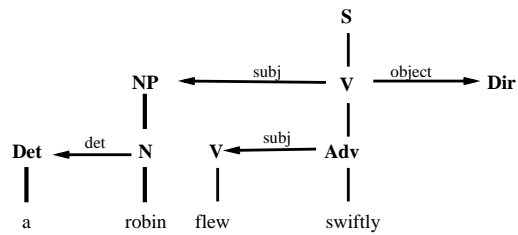,height=32mm} \end{center}
\caption{Graph built before backtracking}
\label{incompleteRobin.fig}
\end{figure}

Given these two rules, the sentence {\em a robin flew swiftly}
would first be mapped to the parse graph on
Figure~\ref{incompleteRobin.fig}; then, on realizing that, by the
end of the sentence, the parse graph is still incomplete, one
would have to reconsider the choices made. Supposing there is a
single analysis available for {\em swiftly}, the parser would
then backtrack on {\em flew} and try the second rule, which would
eventually lead to a successful parse.

Backtracking involves undoing some of the rules applied so
far. This supposes that track is kept of the list of rules
applied, and, more importantly, that each rule is reversible. As
a matter of fact, rules, as defined, {\em are} reversible (owing to
Principle~\ref{newNodes.princ},
Section~\ref{substitutionRevisited.sec}). In order to undo the
effect of a rule, the addendum is to be removed and the
interpolation path replaced back by the context anchor.  If rules
were not reversible, then, it would be necessary to keep track of
at least the last few parse configurations.

Of course, backtracking does not always take place after reading
the whole input string. It takes place whenever no
further rule can be triggered and the parse graph is incomplete.

For another example, consider the sentence {\em he gave
trouble to us} and suppose that the highest-priority rule for
{\em gave} is the following. (The edge label {\em iObj} stands
for ``indirect object''.)

\begin{figure}[htbp]
\begin{center} ~\psfig{file=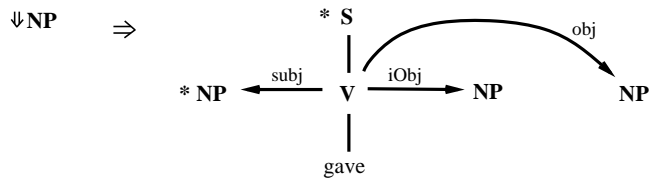,height=25mm} \end{center}
\caption{Default rule for {\em gave}}
\label{giveWithDat.fig}
\end{figure}

Then, on encountering {\em to}, no rule will match the context
(in which an object NP is expected), thus leading the parser to
backtrack on {\em trouble}, then {\em gave}, and select the
following rule, which will eventually lead to a successful
parse. (A {\em to}-Prep is a category that contains only the
preposition {\em to} ---and {\em unto} in archaic dialects.)

\begin{figure}[htbp]
\begin{center} ~\psfig{file=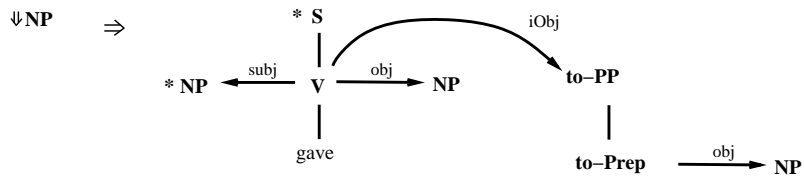,height=25mm} \end{center}
\caption{Alternate rule for {\em gave}}
\label{alternateGave.fig}
\end{figure}

\subsection{Interpolation and anchor matching}
\label{revisit.sec}

This section revisits concepts introduced in
Section~\ref{rule.sec}, giving more precise formulations and
adding details that were initially overlooked.

\subsubsection{Addendum incorporation}
\label{substitutionRevisited.sec}

In all the examples given so far, an insertion consists in
replacing a dangling node with a node with a child and a proper
interpolation consists in interpolating a path whose first edge
is a {\em parent-of} edge, so that the destination of the
interpolation path dominates its source.

With interpolations of this type, merging an addendum into the
context is a fairly straightforward process, which has not been
explicated so far, but only illustrated through examples. 

However, the concept of interpolation allows more flexibility than
has been hitherto suggested, and, in order to consider further
varieties of interpolation, it is important to bring out the graph
invariants implicitly used and preserved during addendum
incorporation.

\begin{invariant}
In a phrase graph, there cannot be distinct edges with the
same label.
\label{distinctFunctions.inv}
\end{invariant}

\begin{invariant}
In a parse graph, any node except the root has exactly one
incoming edge.
\label{oneIncomingEdge.inv}
\end{invariant}

\begin{invariant}
In a parse graph, there cannot be more than one {\em
parent-of} edge from any node.
\label{oneChild.inv}
\end{invariant}

According to invariant~\ref{oneIncomingEdge.inv}, an
interpolation path can be defined as the unique oriented path
that connects two addendum anchors. With respect to this path,
there is a {\em source anchor} and a {\em destination anchor}.
Invariant~\ref{oneIncomingEdge.inv} also implies that the source
anchor inherits the incoming edge of the context anchor as a
result of an interpolation.

Furthermore, when one looks at the way the outgoing edges of the
context anchor are distributed between the addendum anchors of an
interpolation, invariant~\ref{oneChild.inv} can be seen to prevail in
that the {\em parent-of} edge, in the examples seen so far, is
transmitted to the destination anchor. The underlying principle
can be stated as follows.

\begin{principle}
\label{outgoingDistrib.princ}
All edges of the context anchor are inherited by the source
of the interpolation path except for edges present in the
addendum, which are shifted to the destination of the
interpolation path.
\end{principle}

Principle~\ref{outgoingDistrib.princ} itself presupposes that
the addendum and the context form
disjoint graphs. This principle is well worth spelling out, for
it allows rules to be easily reversible in the sense of
Section~\ref{backtracking.sec}.

\begin{principle}
\label{newNodes.princ}
In a composition rule, the sets of nodes of the addendum and the
context have an empty intersection.
\end{principle}

Indeed, if the graph union that is specified by a rule is a
disjoint union, then, no information beyond the list of rules
applied is necessary to undo the effect of these rules.

On the other hand, Principle~\ref{outgoingDistrib.princ}
presupposes that, if the context anchor is a head, then, the
addendum anchors must both be capable of inheriting functional
edges, i.e. be phrase heads as well. Conversely, if the context
anchor is not a head, substituting a head for it would create a
second head within one phrase, which is ruled out by
definition. These considerations can be summarized by the
following principle.

\begin{principle}
\label{headnessMatching.princ}
The head status of the context anchor must be matched by the
addendum anchors.
\end{principle}

This principle does not directly govern addendum incorporation,
for it has to be built into the grammar rules. It is especially
useful as a prerequisite of Principle~\ref{outgoingDistrib.princ}.

\subsubsubsection{Nonstandard interpolation}

\begin{figure}[htbp]
\begin{center} ~\psfig{file=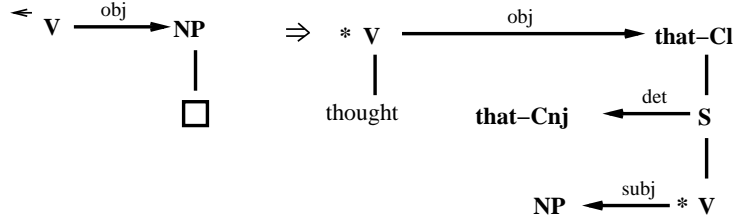,height=32mm} \end{center}
\caption{A nonstandard case of proper interpolation}
\label{nonstandardInterpolation.fig}
\end{figure}

To see how Principle~\ref{outgoingDistrib.princ} applies to a
nonstandard case of proper interpolation, i.e. a case in which
the interpolation path does not start with a {\em parent-of}
edge, consider the rule on
Figure~\ref{nonstandardInterpolation.fig}. (The box represents an
empty category, or gap; the symbol {\em that}-Cnj informally
denotes a subcategory whose main representative is the
conjunction {\em that}.)

Although the context anchor, which is a verb, is a dangling node,
an empty object~NP (i.e. an object {\em gap}) has been predicted,
typically owing to the presence of a preceding object pronoun,
i.e. a relative or interrogative pronoun, or a topicalized
object~NP. The occurrence of a verb that governs a
sentence rather an object NP pushes the predicted object gap
further. Since the object gap does not appear in the addendum
(owing to Principle~\ref{newNodes.princ}), its position in the
graph after the rule has applied is not specified explicitly, but
can be inferred from
Principle~\ref{outgoingDistrib.princ}. Indeed, since the {\em
object} edge from the context anchor is redundant with the {\em
object}~edge that emanates from the source anchor of the
addendum, it is inherited by the destination anchor. Therefore,
after addendum incorporation, the context will contain the
following subgraph.

\begin{figure}[htbp]
\begin{center} ~\psfig{file=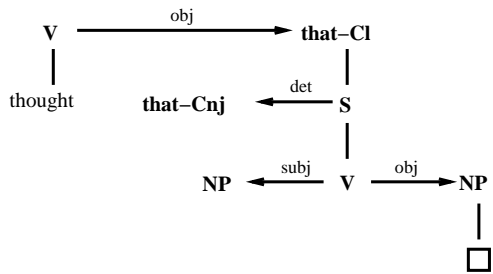,height=37mm} \end{center}
\caption{Subgraph obtained after addendum incorporation}
\label{resultingSubgraph.fig}
\end{figure}

\subsubsubsection{Nonstandard insertion}
\label{nonstandardInsertion.sec}

A standard insertion substitutes a node with an outgoing {\em
parent-of}~edge for a dangling node. But it is also conceivable
to substitute a node with arbitrary nonredundant outgoing edges
for a node with descendants.

A practical use of nonstandard interpolation could be the
handling of optional complements.  For example, supposing the
rule that handles the {\em by}-complement of a passive were the
one represented on Figure{passiveAgent.fig}, there would be no
necessity to anticipate the presence of a {\em by}-complement on
encountering a passive, and so no backtracking would be involved
to handle its optionality\footnote {The directional complement of
the verb {\em fly}, used to illustrate backtracking in
Section~\ref{backtracking.sec}, could be handled in this way,
unless there were good reasons to distinguish the two
constructions from the point of view of lexical semantics.}.

\begin{figure}[htbp]
\begin{center} ~\psfig{file=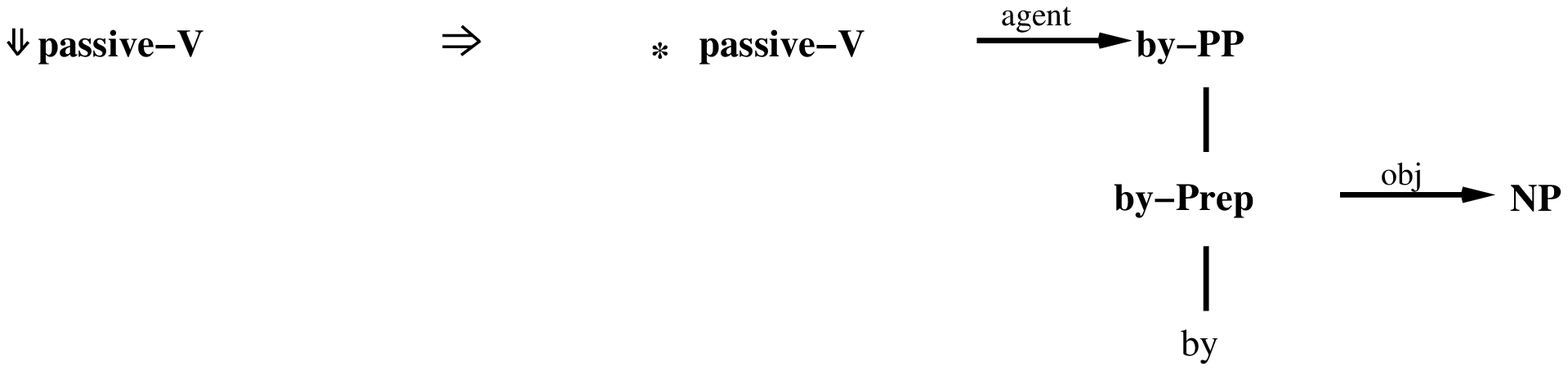,height=33mm} \end{center}
\caption{Rule for the {\em by}-complement of a passive}
\label{passiveAgent.fig}
\end{figure}

\subsubsection{Anchor matching}
\label{anchorMatching.sec}

The principles assumed so far for matching the context anchor
have to be stated explicitly and possibly extended; this is
necessary, on the one hand, to predict the anchor position for
any arbitrary interpolation, including a nonstandard one, and, on
the other hand, to cater for variations in word order.

Section~\ref{basicAnchorMatching.sec} specified two~possible
positions for the context anchor, as ancestor or immediate linear
successor of the previous lexeme. But no principle to find out
whether a given interpolation should take place at an ancestor or
successor position has yet been stated. As a matter of fact, a
simple principle underlies the examples given so far.

\begin{principle}
\label{anchorChoice.princ}
If the addendum lexeme linearly follows either endpoint of
the interpolation path, the context anchor is an ancestor of the
previous lexeme. In all other cases, the context anchor is the
immediate linear successor of the previous lexeme.
\end{principle}

This principle is analogous to
Principle~\ref{ancestorMatching.princ} in that it prevents the
occurrence of dangling nodes among the linear predecessors of the
current lexeme. So it is a practical consequence of a more
general principle.

\begin{principle}
\label{barDanglingNodes.princ}
A valid interpolation cannot leave or create dangling nodes that
linearly precede the lexeme it integrates.
\end{principle}

Another consequence of Principle~\ref{barDanglingNodes.princ} on
the form of rules is the following.

\begin{principle}
\label{tightAddendum.princ}
An addendum cannot contain dangling nodes that linearly precede
the anchor that inherits the {\em parent-of} edge of the context
anchor. 
\end{principle}

Under strict word order, the form of rules can be controlled
through Principles~\ref{tightAddendum.princ}
and~\ref{anchorChoice.princ} to enforce
Principle~\ref{barDanglingNodes.princ}. When alternative
orderings are allowed, the matching process itself may have to
take this principle into account, for rules may have to admit
addenda that {\em may} violate
Principle~\ref{barDanglingNodes.princ}
or~\ref{tightAddendum.princ} under conditions that can only be
checked dynamically.

\begin{figure}[htbp]
\begin{center} ~\psfig{file=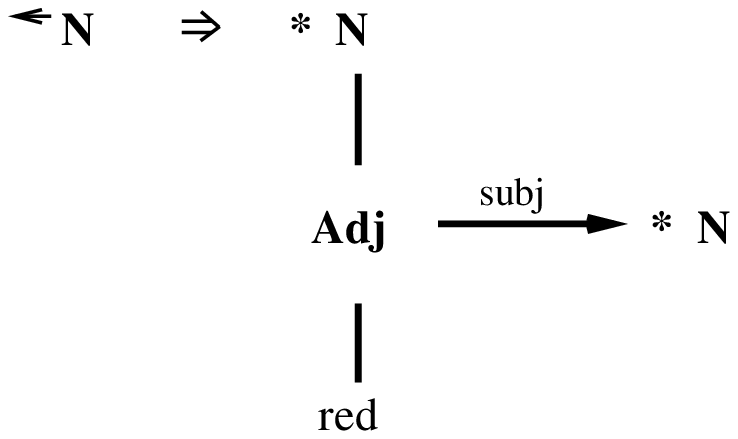,height=28mm} \end{center}
\caption{Premodification by the adjective {\em red}}
\label{premod.fig}
\end{figure}

\begin{figure}[htbp]
\begin{center} ~\psfig{file=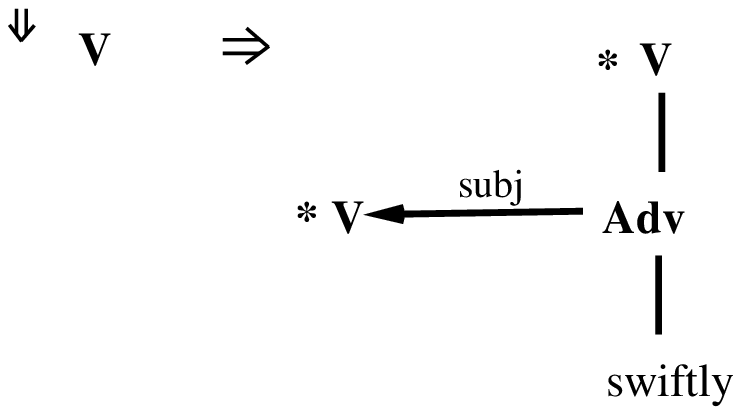,height=28mm} \end{center}
\caption{Postmodification by the adverb {\em swiftly}}
\label{postmod.fig}
\end{figure}

To see Principle~\ref{anchorChoice.princ} in application, compare
the rule on Figure~\ref{premod.fig}, which represents a
premodification, to the rule on Figure~\ref{postmod.fig}, which
represents a postmodification.

It is now apparent that word order plays a key role in anchor
choice during rule design and in anchor matching during rule
application. So it is interesting to see what extensions to the
principles just brought out are necessary in order to accommodate
greater freedom in word order than has been presupposed up till
now.

\subsubsubsection{Complex phrase ordering}
\label{freeWordOrder.sec}

Under a strict ordering of phrase nodes, a parse graph can be
linearized by considering the frontier of the summary tree
(Definition~\ref{counterpartTree.def},
Section~\ref{parseGraphs.sec}). Then, in a given parse
configuration, the previous lexeme has a unique immediate
successor. However, when several possible orders are allowed,
there may be more than one linear successor to the previous
lexeme. 

Consequently, several context nodes may be found to match the
context anchor of an applicable rule. This introduces a degree of
indeterminacy which is compounded with the indeterminacy
resulting from rule conflicts (i.e. the availability of several
matching rules for a given lexeme).

Not only does this type of indeterminacy arise in the
presence of free word order, but it is also liable to arise
in a fixed-order framework if several nodes along a {\em
parent-of} path are allowed to bear the same label. Indeed, this
situation could result in several nodes matching an ancestor
anchor; and a principle that ruled it out would be no more than an
{\em ad hoc} feature that reduced the formalism's expressive power
without reducing its indeterminacy in any essential
way. Therefore, indeterminacy due to anchor matching has to be
handled somehow.

Once again, a trial-and-error approach seems to be the natural
path to follow. The mechanism used to solve rule conflicts can be
applied here too, provided competing anchor matches are somehow
ordered by priority. As far as ancestor conflicts are concerned,
proximity to the previous lexeme (i.e. depth) seems to correlate
simply and naturally with priority. As far as successor conflict
is concerned, however, it would seem that no natural solution is
available unless, even when the placement of nodes in a phrase
graph is relatively free, there is an underlying preferred order.

In addition to the possibility of finding multiple anchor
matches, free word order also complicates the process of
identifying context anchors, for context nodes do not stand in a
simple topological relation with respect to the previous
lexeme. From an algorithmic point of view, the solution seems to
consist in updating the set of successors of the previous lexeme
after each interpolation, as described in
Procedure~\ref{updateAnchorSet.proc},
Appendix~\ref{locateAnchor.app}.

\subsection{Coreference links}
\label{coref.sec}

Incremental left-to-right parsing, as outlined, forms phrases
exclusively from adjacent lexemes. However, immediate syntactic
relations between distant constituents do exist in natural
languages. Interrogative and relative clauses in English provide
such examples. For example, in the following sentence, the
pronoun {\em who} somehow functions as the subject of {\em to
win}, although these two constituents occupy distant positions in
the input stream.

\begin{quote}
Who do you want to win?
\end{quote}

This difficulty will be solved through a now classic device in
syntactic theory, namely coindexing of phrases. This consists in
linking coreferential phrases together, typically by assigning to
them equal referent indices. In the above example, the
interrogative pronoun {\em who} and the empty subject of {\em to
win} (whose presence is concretized by the unacceptability of
{\em Who d'you wanna win?} as a spoken realization) are
coindexed.

In fact, in order to reconcile the fact that the parsing model
under discussion can only establish immediate syntactic relations
between neighbouring constituents and the fact that various
ordering constraints or rhetorical phenomena may separate phrase
constituents, we will rely on coreference links. A coreference
link is somewhat more general than {\em NP}~coindexing is that
it may apply to any part of speech rather than just noun phrases.

Nodes related by a coreference links have just one counterpart in
the semantic interpretation to be derived from the parse
graph. This means that all members of a coreference chain denote
the same object in the universe of discourse. Syntactically, this
entails that members of a coreference chain of NPs share gender
and number characteristics, and members of a coreference chain of
verbs share complements. (For example, one member will have a
subject, another member no subject but a direct object, and a
third an indirect object, thus forming three instances of a
single verb with subject, object, and indirect object.)

Many phenomena in natural languages suggest that the human parser
prohibits to a large extent the splitting of a phrase across
distant lexemes but uses coindexing to handle the sharing of
constituents between phrases and rhetorical disruptions to
standard phrasal order.

Such disruptions are typically attenuated by the use of
resumptive pronouns in spoken French, as in the following
example.

\begin{quote}
Il est bizarre, ton chapeau. \\
\it It is weird, [is] your hat.
\end{quote}

Here, the main verb finds its subject where the parser expects
it, but this subject is only a pronoun coreferenced with the
``real'' subject, which, for intonational effect, is relegated to
a secondary tone unit so as to allow full emphasis on the
predicate.

When no overt pronoun shows the existence of a coreference chain,
it is sometimes necessary to postulate the existence of empty
lexemes, as in the interrogative clause cited above.

In Section~\ref{DCSD.sec}, an analysis of Dutch cross serial
dependencies will be attempted in terms of covert coreference
chains.

The introduction of coreference links gives rise to the following
definitions.

\begin{definition}
\label{apg.def}
An {\em augmented parse graph} is a parse graph augmented with
coreference links.
\end{definition}

\begin{definition}
\label{corefLink.def}
A {\em coreference link} is an undirected edge that relates two
nodes standing in an equivalence relation. Links that can be
inferred, such as the link from any node to itself, do not need
to appear in the representation of an augmented parse graph.
The equivalence relations represented in an augmented parse graph are
{\em coreferentiality relations}.
\end{definition}

\begin{definition}
\label{coreferentiality.def}
A {\em coreferentiality} is an equivalence relation over parse graph
nodes bearing a specified label. Coreferentialities found useful
are NP~coreferentiality and verb~coreferentiality.
\end{definition}

This defines a graph hierarchy, with phrase graphs at the bottom,
and augmented parse graphs at the top.

\subsection{Multiple interpolation}
\label{multipleSubstitution.sec}

\begin{figure}[htbp]
\begin{center} ~\psfig{file=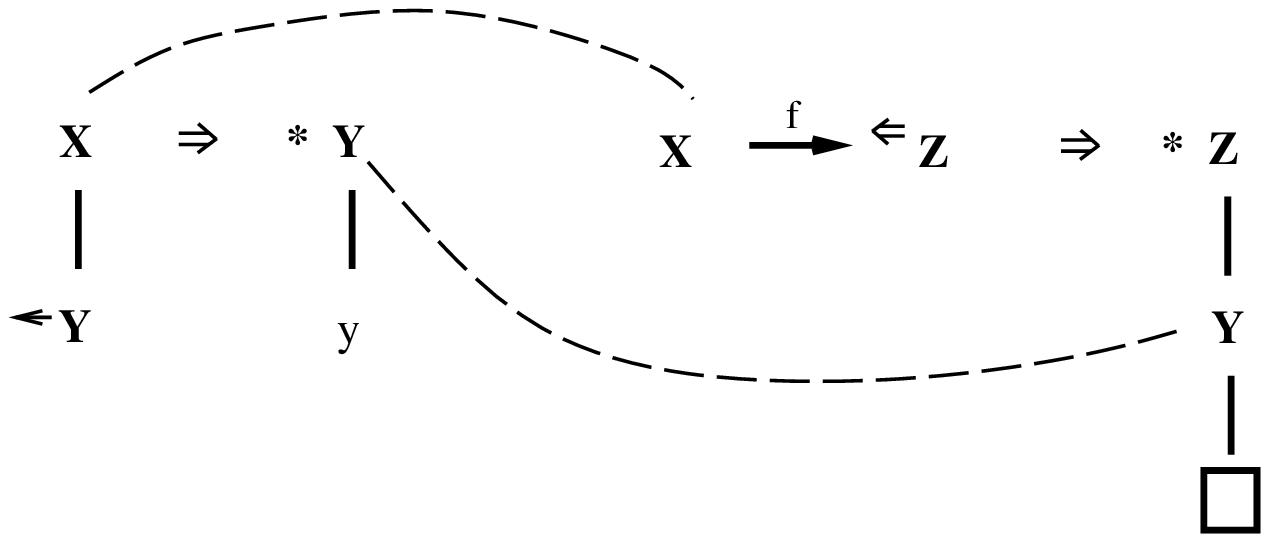,height=3cm} \end{center}
\caption{A double interpolation}
\label{doubleSubst.fig}
\end{figure}

Coreference links make it meaningful to define parallel
interpolations operating simultaneously on several context
anchors. Multiple interpolation seems useful only when the
context anchors are located with respect to a coreference
chain. Figure~\ref{doubleSubst.fig} shows an example of a double
interpolation.

This example is purely formal, but concrete examples will be given
in Section~\ref{DCSD.sec} on Dutch Cross-Serial
Dependencies. Nonetheless, it illustrates an important feature of
multiple interpolation; namely, the fact that the rule, albeit
multiple, adds a single lexeme to the parse graph. Typically, it
anticipates a $Y$-gap coreferential with this lexeme. Gaps do not
count as lexemes in so far as they are not physically present in
the input string.

Furthermore, the formulation of multiple interpolation makes it
necessary to define a third possible location for the context
anchor, namely {\em linear successor}, denoted by a double left
arrow (\successor), as opposed to {\em immediate linear
successor}, denoted by a simple left arrow (\immediateSuccessor). 

\subsection{Implicit reflexive-transitive closure in patterns}
\label{KleeneStar.sec}

The definition of coreferentiality as an equivalence relation
(Definition~\ref{coreferentiality.def}) yields somewhat
unsuspected power to the occurrence of coreferential links in
context patterns.

\begin{figure}[htbp]
\begin{center} ~\psfig{file=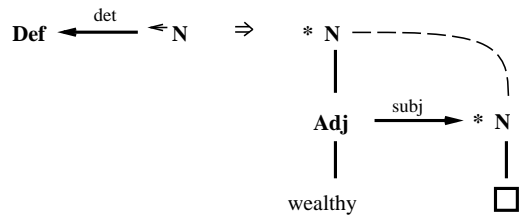,height=3cm} \end{center}
\caption{Restrictive rule for adjectival definite descriptions}
\label{tentativeNounless.fig}
\end{figure}

To see that, consider the example of adjectival definite
descriptions in English, as found in {\em the wealthy}, {\em the
influential}, etc. Leaving semantic features aside, one important
syntactic requirement for forming such NPs is that the determiner
be definite. This is captured by the rule on
Figure~\ref{tentativeNounless.fig}. ({\em Def} stands for ``definite
determiner''.)

Coreference links are created because the structure being built
is a modification structure, not because of the presence of a
noun gap\footnote {This ``gap'' does not play a pronominal role,
but rather represents a semantically empty noun. A precise
notation would include the name {\em Pro} in the label of
``real'' gaps, such as those used in Section~\ref{DCSD.sec}.}.
The modification examples given so far did not contain
coreference links simply because extended parse graphs had not
been defined.

\begin{figure}[htbp]
\begin{center} ~\psfig{file=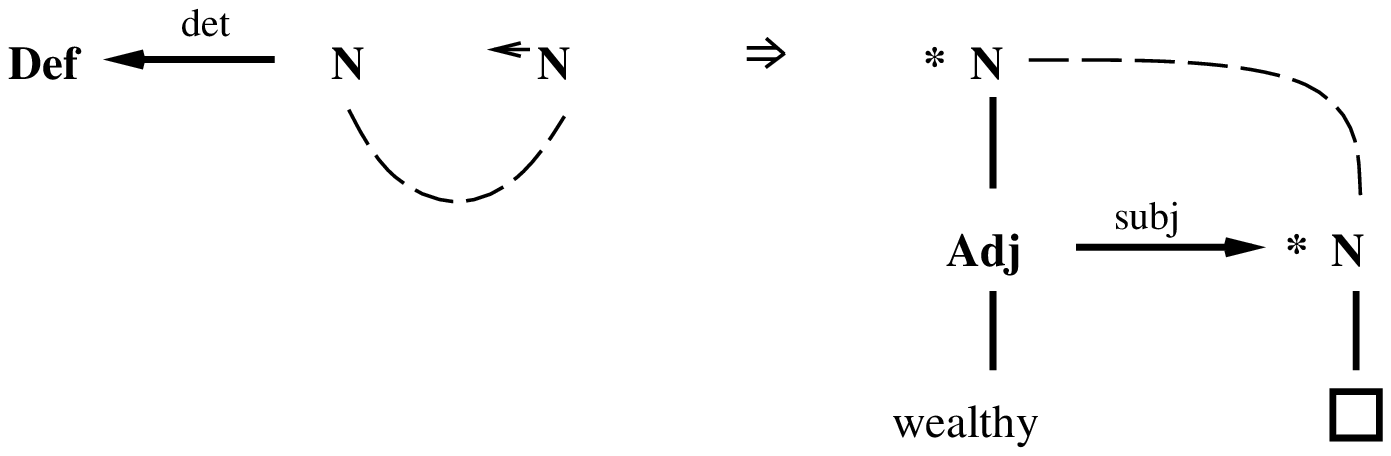,height=3cm} \end{center}
\caption{Comprehensive rule for adjectival definite descriptions}
\label{nounless.fig}
\end{figure}

This rule requires that the noun node that is to be replaced by
an adjective and an empty noun be immediately preceded by a
definite determiner. Therefore, it will not apply to a noun which
is itself premodified by an adjective, as in {\em the idle
rich}. On the other hand, one cannot devise a rule covering
specifically a string of the form {\em Def~Adj~Adj}, for the
premodification can be iterated, as in {\em the outrageous idle
rich}. Consider therefore the rule on Figure~\ref{nounless.fig}.

\begin{figure}[htbp]
\begin{center} ~\psfig{file=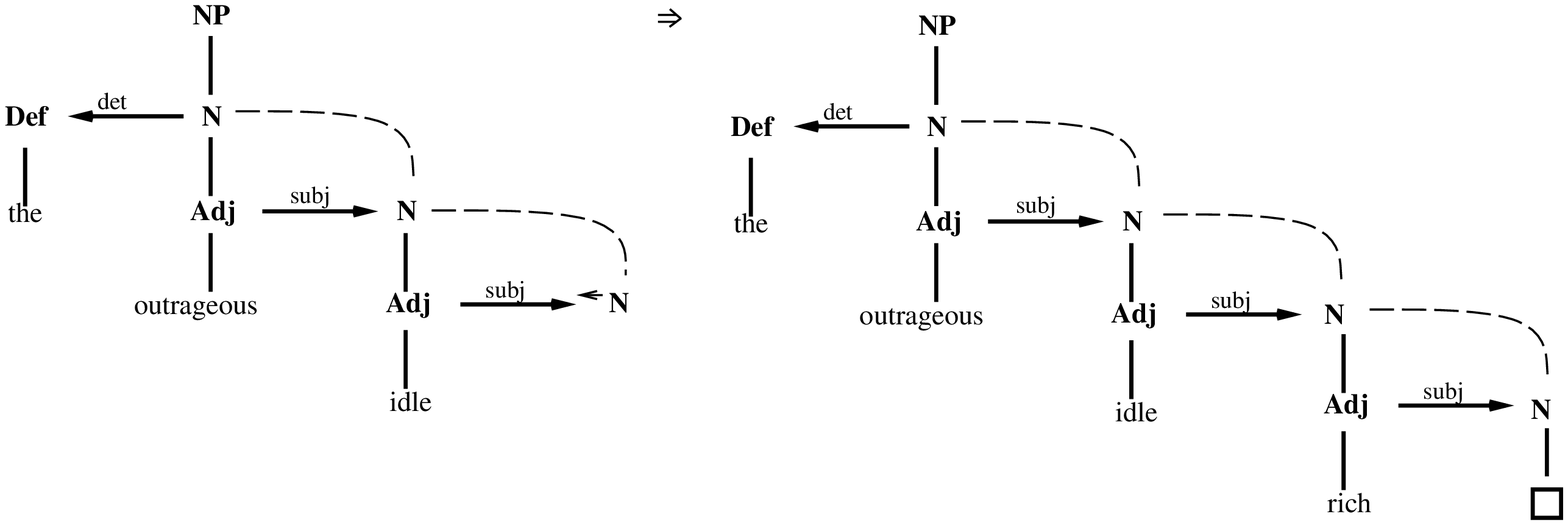,width=\textwidth} \end{center}
\caption{Last interpolation performed when parsing {\em the outrageous idle rich}}
\label{outrageous.fig}
\end{figure}

This rule allows the context anchor to be separated from the
NP~head by any number of premodification stages, including zero.
For example, the last interpolation to be applied when parsing
{\em the outrageous idle rich} can be represented by the
diagram on~\ref{outrageous.fig}\footnote
{This diagram represents a derivation step, as explained in Section~\ref{sum.sec}.}.

The existence of rules in which an anchor is related to the rest
of the context by a coreference link leads to the following
definition of a context pattern.

\begin{definition}
\label{contextPattern.def}
A {\em context pattern} is a subgraph of an augmented parse graph
with one or several distinguished nodes called {\em
anchors}. Anchor location with respect to the previous lexeme
takes one of the three values {\em ancestor}, {\em immediate
successor}, and {\em successor}.
\end{definition}


\section{A simple expression grammar}
\label{cfg.sec}

To show the specificities of \GIGs, even in the absence of
context-sensitivity in rules, this section applies them to the
classical problem of a simple arithmetic expression language with
two precedence levels and left associativity.

\subsection{Deriving an GIG from a CFG}

This section will derive as simply as possible a \GIG\ from the
Context Free Grammar given in Figure~\ref{cfg.fig}. This kind of
derivation could conceivably be performed automatically in order
to benefit from the incrementality of GIG-driven parsing for
existing context-free grammars. For example, in a language-based
editor, the incremental building of parse representations as each
token is read could be used to implement syntax-sensitive editing
functions.

\begin{figure}[htbp]
\begin{tabular}{lll}
E & $\rightarrow$ & E $\bf +$ T \\
E & $\rightarrow$ & T \\
T & $\rightarrow$ & T {\bf *} F \\
T & $\rightarrow$ & F \\
F & $\rightarrow$ & {\bf (} E {\bf )} \\
F & $\rightarrow$ & {\bf 0} \\
F & $\rightarrow$ & {\bf 1} \\
\end{tabular}

\caption{Context-Free Grammar for a simple expression language}
\label{cfg.fig}
\end{figure}

\subsubsection{Numbers}

\begin{figure}[htbp]
\begin{center} ~\psfig{file=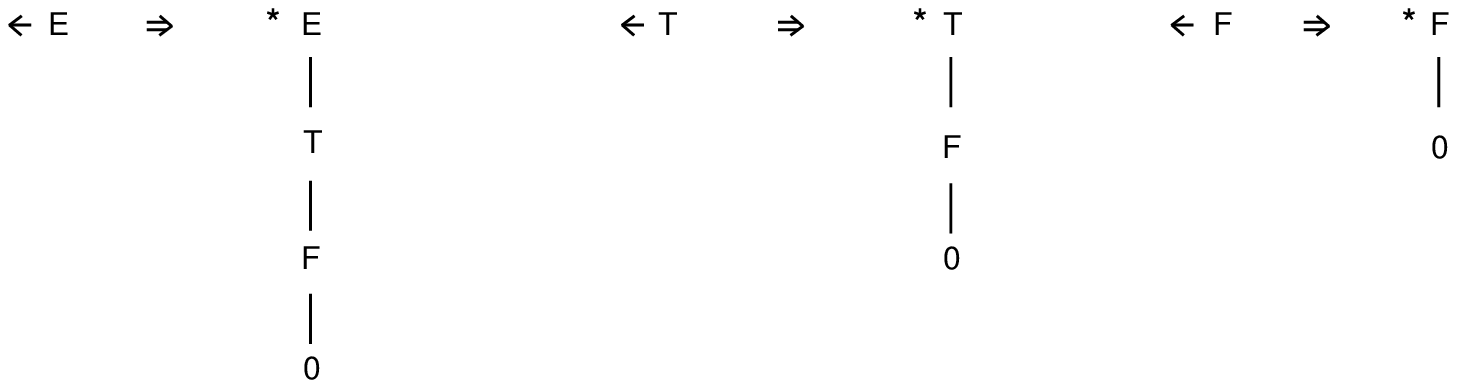,height=30mm} \end{center}
\caption{GIG rules associated with the lexeme {\em 0}}
\label{zero.fig}
\end{figure}

Moving from a CFG to a GIG involves lexicalization. This means
that every rule is to be matched by a lexeme and contains a representation of the
contexts in which this lexeme is likely to occur. As far as
numbers are concerned, they can function either as factors, terms,
or expressions. Therefore, each number will have three
associated rules. Figure~\ref{zero.fig} shows the three~rules
associated with the number~0.

\subsubsection{Sum}
\label{sum.sec}

\begin{figure}[htbp]
\begin{center} ~\psfig{file=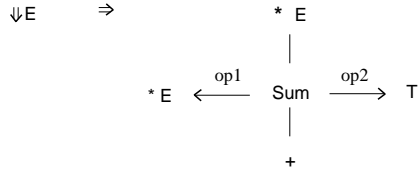,height=23mm} \end{center}
\caption{Rule to be matched on encountering a $+$}
\label{sum.fig}
\end{figure}

A sum is identified on scanning the lexeme~$+$. By representing a
sum as an interpolation into an~{\em E}, the fact that $+$~is
the lowest-priority operator is captured. Accordingly, the rule
for $+$ is as represented on Figure~\ref{sum.fig}.

Since the addendum lexeme follows the destination of the
interpolation path, i.e. the first operand of the sum, the
context anchor is stipulated to be an ancestor of the previous
lexeme (Principle~\ref{anchorChoice.princ}).

\begin{figure}[htbp]
\begin{center} ~\psfig{file=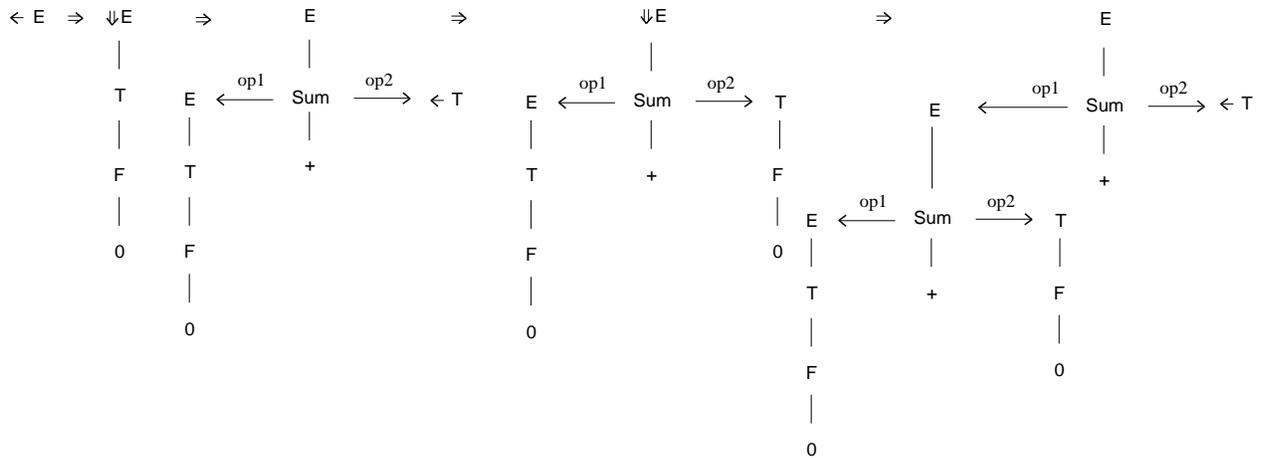,width=\textwidth} \end{center}
\caption{The derivation for $0 + 0 +$}
\label{plusDerivation.fig}
\end{figure}

The fact that the second operand is a term ({\em T}) rather than
an expression ({\em E}) forces left associativity. This can be
verified by unwinding the GIG derivation for the string $0 +
0 + 0$. This derivation, up to the insertion of the last zero, is
represented on Figure~\ref{plusDerivation.fig}.

A step in a GIG~derivation consists in the marking of a context
anchor in the result of the preceding step, and a representation
of the parse graph as modified by the interpolation applied.

\subsubsection{Product}

The rule for a product is entirely analogous to the rule for a
sum, except that it is allowed to apply to a term rather than an
expression. This fact gives it precedence over sum, for
essentially the same reason as in the original context-free
grammar. 

\begin{figure}[htbp]
\begin{center} ~\psfig{file=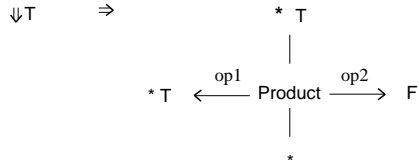,height=23mm} \end{center}
\caption{The rule for $*$}
\label{product.fig}
\end{figure}

\begin{figure}[htbp]
\begin{center} ~\psfig{file=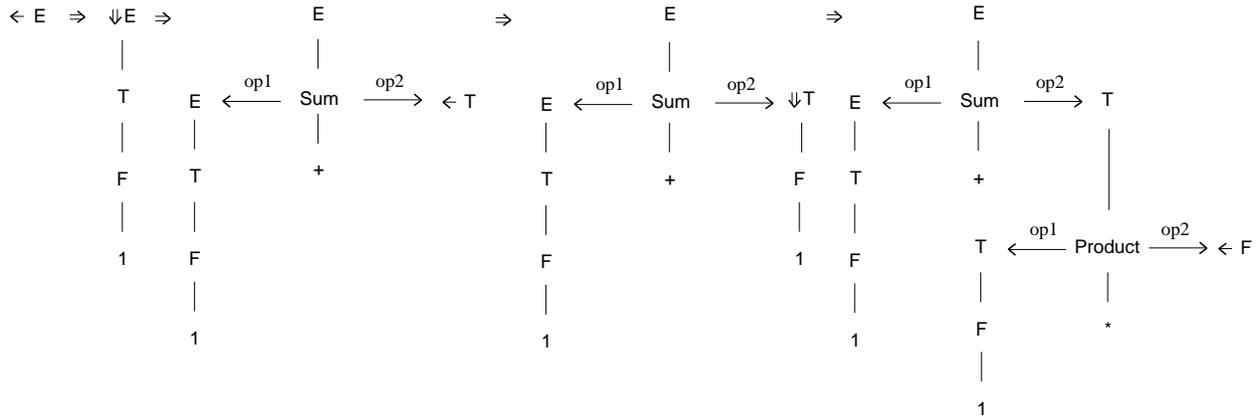,width=\textwidth} \end{center}
\caption{The derivation for $1 + 1 *$}
\label{productDerivation.fig}
\end{figure}

Figure~\ref{product.fig} shows the rule for the product operator,
and Figure~\ref{productDerivation.fig} shows the derivation for
$1 + 1 * 0$ up to the insertion of the lexeme~$0$. It
illustrates the relative priorities of sum and product.

\subsubsection{Parentheses}

\begin{figure}[htbp]
\begin{center} ~\psfig{file=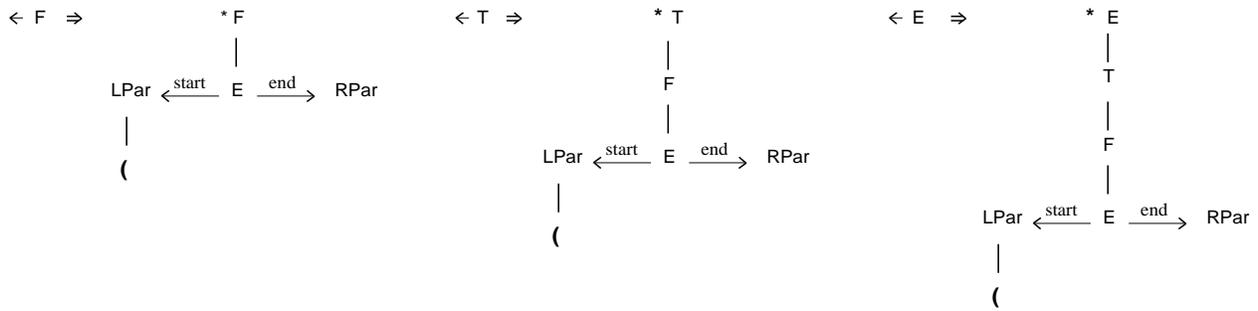,width=\textwidth} \end{center}
\caption{The rules for the left parenthesis}
\label{lParen.fig}
\end{figure}

A rule for a left parenthesis is an insertion that announces an
expression and a right parenthesis in a context where an
expression, a term, or a factor could be expected. As a right
parenthesis is not supposed to occur unless it has been announced
by a left parenthesis, its rule is a simple lexical
insertion. Accordingly, only the rules for the left parenthesis are
shown on Figure~\ref{lParen.fig}.

Note that nonterminals {\em RPar} and {\em Rpar} are necessary
according to principle~\ref{lexicalNode.princ},
Section~\ref{parseGraphs.sec}.

\begin{figure}[htbp]
\begin{center} ~\psfig{file=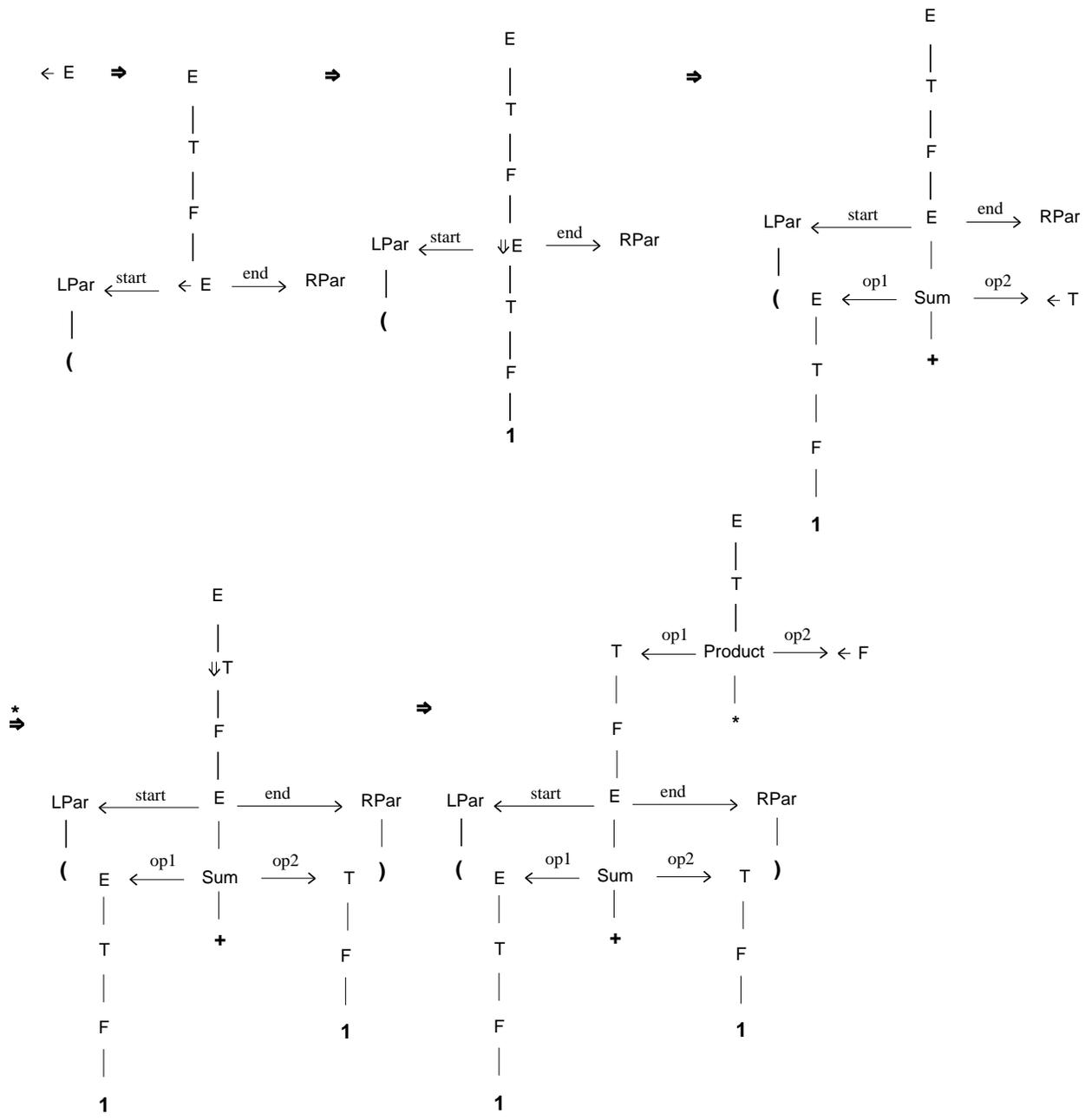,width=\textwidth} \end{center}
\caption{Main steps in the derivation for $(1 + 1) * 0$}
\label{lParenDerivation.fig}
\end{figure}

Applying these rules to the parsing of $(1 + 1) * 0$, one
obtains the derivation whose main steps are shown on
Figure~\ref{lParenDerivation.fig}.

\subsection{A more idiomatic approach}
\label{expressionWithInheritance.sec}

The existence of three rules to insert a number or a left
parenthesis is the translation of a CFG idiom for handling
precedence. The native GIG device for solving this type of
problem relies on node subtyping.

\begin{figure}[htbp]
\begin{center} ~\psfig{file=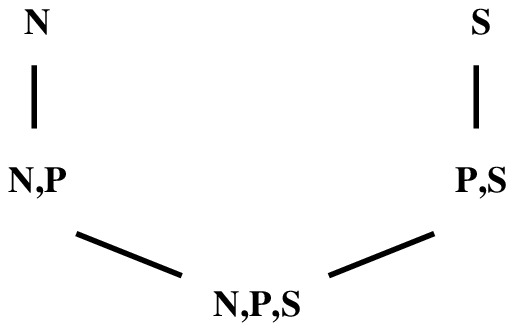,height=2cm} \end{center}
\caption{Type hierarchy for the expression grammar}
\label{typeHierarchies.fig}
\end{figure}

For the expression language considered, we can have tree~type
atoms, {\em N} (for {\em number}), {\em P} (for {\em product}),
and {\em S} (for {\em sum}). They form two hierarchies that meet
at the universal type {\em N,P,S}, as shown on
Figure~\ref{typeHierarchies.fig}.

If we ignore preterminal labels (such as {\em Sum} or {\em
LPar}), an {\em N,P,S} matches any anchor; so it is an expression
in a broad sense. It occurs typically as the initial parse graph
and between parentheses.  An {\em N,P} is any expression except a
sum, so it occurs as the right operand of a sum to enforce left
association. Likewise, an {\em N} is exclusively a number (or a
parenthesized expression), such as occurs as the right operand of
a product. As far as the right-hand branch of the hierarchy is
concerned, a {\em P,S} can occur as the left operand of a product
or sum, while an {\em S} can occur only as the left operand of a
sum. These enforce the relative precedence of sum and product.

\subsubsection{Numbers}

\begin{figure}[htbp]
\begin{center} ~\psfig{file=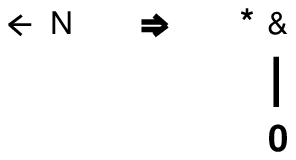,height=15mm} \end{center}
\caption{The rule for zero}
\label{zeroBis.fig}
\end{figure}

There is a single rule for inserting a number. It requires the
atom {\em N} in the context anchor and does not modify its type.
The symbol~{\em \&} that is used as the label of the addendum
anchor indicates that this label is inherited as is from the node
matching the context anchor. The rule for zero is represented on
figure~\ref{zeroBis.fig}.

\subsubsection{Sum}

\begin{figure}[htbp]
\begin{center} ~\psfig{file=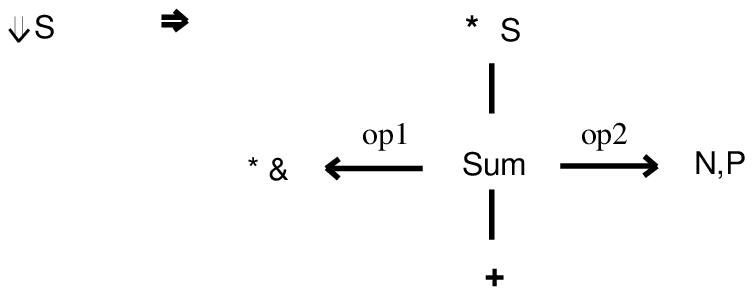,height=23mm} \end{center}
\caption{The rule for $+$}
\label{plusBis.fig}
\end{figure}

The rule for the operator~$+$ is represented on
Figure~\ref{plusBis.fig}.

A sum can be interpolated only at a node whose type contains the
type atom {\em S}. To guard against the interpolation of a
product on top of a sum, the type {\em S} is assigned to the root
of the addendum. To guard against right association of additions,
the type {\em N,P} is assigned to the right operand.  The left
operand cannot be the object of further interpolations and simply
inherits the type of the node matching the context anchor.

\begin{figure}[htbp]
\begin{center} ~\psfig{file=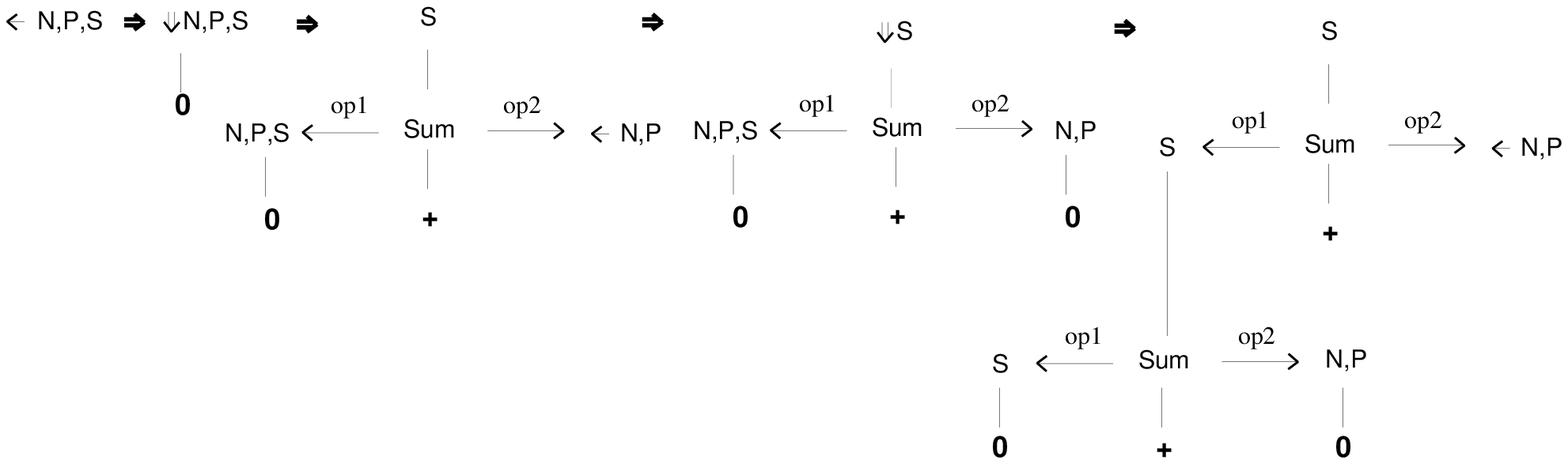,width=\textwidth} \end{center}
\caption{The derivation for $0 + 0 +$}
\label{plusBisDerivation.fig}
\end{figure}

Figure~\ref{plusBisDerivation.fig} derives $0 + 0 + 0$ up to
the last insertion to show that left associativity is enforced.

When the second occurrence of $+$ is encountered, only one of
the ancestors of the previous lexeme has a type which contains
{\em S}.

\subsubsection{Product}

\begin{figure}[htbp]
\begin{center} ~\psfig{file=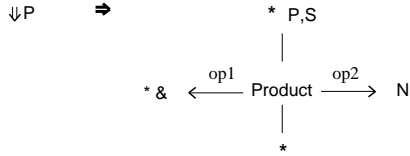,height=23mm} \end{center}
\caption{The rule for the product operator}
\label{productBis.fig}
\end{figure}

The rule for the operator~$*$ exactly parallels the rule for
the sum operator. It requires the anchor type to contain the atom
{\em P}, assigns the type {\em P,S} to the addendum root, and
assigns the type {\em N} to the right operand. It is represented
on Figure~\ref{productBis.fig}.

\begin{figure}[htbp]
\begin{center} ~\psfig{file=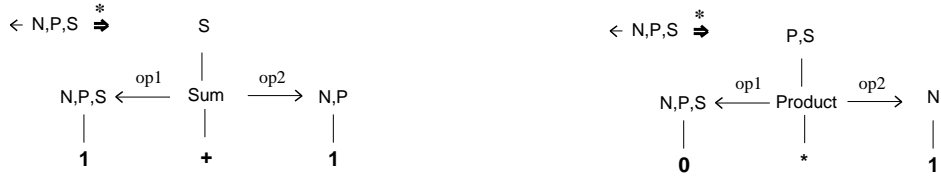,height=24mm} \end{center}
\caption{Derivation outlines for $1 + 1$ and $0 * 1$}
\label{productBisDerivations.fig}
\end{figure}

The fact that the root node of the addendum is a {\em P,S} rather
than an {\em S} gives priority to multiplication over addition.
Figure~\ref{productBisDerivations.fig} outlines the derivations
for $1 + 1$ and $0 * 1$. 

If the current lexeme after parsing $1 + 1$ is~$*$, only
the closest ancestor of the previous lexeme will have a matching
type. On the other hand, if the current lexeme after parsing $0 *
1$ is a plus, only the root of the parse graph will have a
matching type.

\subsubsection{Parentheses}

\begin{figure}[htbp]
\begin{center} ~\psfig{file=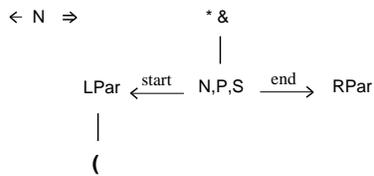,height=23mm} \end{center}
\caption{The rule for the left parenthesis}
\label{lParenBis.fig}
\end{figure}

The rule for the left parenthesis is given on
Figure~\ref{lParenBis.fig}. 

Once it has applied, only an insertion at the linear successor of
the left parenthesis is possible (due to
Principle~\ref{ancestorMatching.princ}). This could be the
insertion of a number or a left parenthesis. So, the supertype
{\em N} as phrase head would also work, but no particular
constraint is necessary here, which is why the most permissive
type is indicated. The same remark holds for the initial
context. In fact, the type {\em N,P,S} makes the grammar more
open to future evolutions, but could be replaced by {\em N}
without impairing its operation.

\section{Dutch Cross-Serial Dependencies}
\label{DCSD.sec}

In Dutch, a few conjunctions, such as {\em dat} ({\em that}) or
{\em omdat} ({\em because}), introduce a clause in which the verb
comes after its object~NP. Now, with perceptual and causative
verbs, i.e. verbs which syntactically take both an NP object and
an infinitive clause object, a double order prevails: all object
NPs occur before the finite verb, in order of increasing nesting
level, as in German for instance; and infinitive verbs occur
after the finite verb in order of increasing nesting level, which
reflects a right-branching construction for clausal objects,
quite unlike German, in which the verbs occur in order of
decreasing nesting level, thus inviting a recursive analysis.

In several papers on DCSDs, examples revolve around hippopotami.
The example given below, taken from~\cite{Rentier94}, does not
depart from this tradition. The indices relate each verb to its
preceding NP~object.

\begin{quote}
\tt ... dat ik Henk$_1$ haar$_2$ de\mbox{ } nijlpaarden$_3$\mbox{ } zag$_1$ helpen$_2$ voeren$_3$ \\
\tt ... that I Henk$_1$\mbox{ } her$_2$ the hippopotamus$_3$ saw$_1$ help$_2$\mbox{ }\mbox{ } feed$_3$ \\
\rm ``that I saw Henk help her feed the hippopotamus''
\end{quote}

The analysis that will be attempted is based on the principle
that, when a verb in a {\em dat}-clause has an NP object and a
clause object, the NP object is constrained to occur before the
verb, and the clause object is constrained to occur immediately
after it.

In itself, however, the existence of different positional
requirements on the NP object and the clausal indirect object
would not seem to require a special syntactic device when the
ditransitive verb governs an ordinary transitive
verb. Thus, in the light of the principle just outlined, one could
expect the following incorrect construction to occur.

\begin{quote}
* ... dat ik haar zag de nilparden voeren.
\end{quote}

In fact, there seems to be an additional constraint that amounts
to forbidding any object in a {\em dat}-clause to occur before
the finite verb. So the correct construction is instead as
follows. 

\begin{quote}
 ... dat ik haar$_1$ de nilparden$_2$ zag$_1$ voeren$_2$.
\end{quote}

To recapitulate, the constraints at work to produce
this construction are the following.
\begin{enumerate}
\item The NP object of any verb in a {\em dat}-clause or a clause
nested therein occurs before it to form an 'inverted' verb
phrase. 
\item The clausal indirect object of a perceptual or causative
verb, which is an infinitive clause with a subject gap, occurs
after it.
\item In a {\em dat}-clause, no direct NP object, whatever the
nesting level at which it occurs, occurs before the finite verb. 
\end{enumerate}
Now, the conjunction of the third constraint with the existence
of a double order on the construction of perceptual and causative
verbs in {\em dat}-clauses produces a rift in the construction,
as a result of which a ditransitive verb may be separated from
its preceding direct object by a theoretically arbitrary
distance. In order to govern two complements across a rift, a
verb, in the general framework adopted here, is in fact realized
as two graph nodes, one of which is a verb gap. To see that, the
representation below shows the phenomenon with one level of
nesting. Verb gaps are represented by square boxes.

\begin{quote}
... dat ik haar $\Box _1$ de nijlpaarden $\Box _2$ zag$_1$ voeren$_2$ 
\end{quote}

This representation postulates that {\em haar} and {\em de
nijlpaarden} occur in inverted VPs headed by verb gaps and that
the verb gaps are coindexed with lexical verbs occurring in a
right-branching construction governed by the finite verb.

If extended one additional nesting level, the representation that
is obtained is the following.

\begin{quote}
... dat ik Henk $\Box _1$ haar $\Box _2$ de nijlpaarden $\Box _3$ zag$_1$ helpen$_2$ voeren$_3$ 
\end{quote}

Furthermore, each non-finite lexical verb heads a clause whose
subject is a gap. Each such NP~gap is coindexed with the
NP~object of the superordinate verb. The coreference pattern for
NPs is the following.

\begin{quote}
... dat ik Henk$_1$ haar$_2$  de nijlpaarden zag $\Box _1$ helpen $\Box _2$ voeren
\end{quote}

In order to capture these hypotheses as GIG rules, one will have
to consider three cases: {\em (i)}~a {\em dat}-clause headed by a
perception or causative verb, {\em (ii)}~a non-finite clause with
an inverted VP whose head is a perception or causative verb, and
{\em (iii)}~a non-finite clause with an inverted VP whose head is
a monotransitive verb.

\subsection{Complex {\em dat}-clause}

\begin{figure}[htbp]
\begin{center} ~\psfig{file=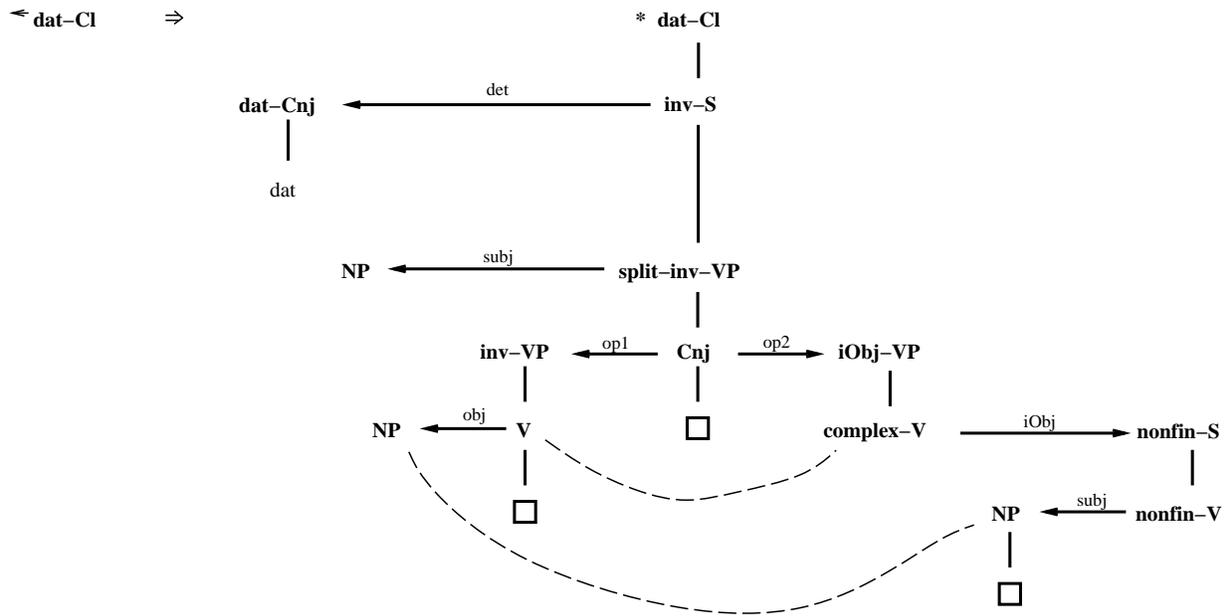,width=\textwidth} \end{center}
\caption{The rule for a {\em dat}-clause with a complex ditransitive}
\label{complexDatClause.fig}
\end{figure}

When the conjunction {\em dat} is encountered, one generally has
no way of guessing that its verb will impose a cross-serial
construction. So the rule on Figure~\ref{complexDatClause.fig}
will be used only on backtracking after realizing that no rule
can integrate the second object~NP {\em haar} if the verb that is
expected is an ordinary transitive verb.

The inverted verb phrase ({\em split-inv-VP}) that heads the sentence is split into an
instance that governs the direct object ({\em inv-VP}) and an instance that
governs the clausal indirect object ({\em iObj-VP}). Since neither of these
half-instances qualifies as a head, an empty coordination
conjunction is postulated to relate them. This
empty conjunction represents the barrier between the
left-branching construction for NPs and the right-branching
construction for verbs.

Given our example, the lexemes whose occurrence is anticipated by
this rule are {\em (i)}~the subject of the {\em split-inv-VP},
namely {\em ik}, {\em (ii)}~the direct object of its {\em inv-VP}
instance, namely {\em Henk}, {\em (iii)}~the {\em complex-V},
namely {\em zag}, and {\em (iv)}~the verb of its clausal object,
namely {\em helpen}, which is labelled as {\em nonfin-V}.

The labels given to nodes here are more precise than required by
the operation of the rules. But this precision hopefully helps
legibility.

\subsection{Complex nonfinite clause}

Just as in the case of the previous rule, the decision to trigger
the rule for an embedded complex verb cannot be taken the first
time its object is encountered, but only after backtracking on
the following~NP.

\begin{figure}[htbp]
\begin{center} ~\psfig{file=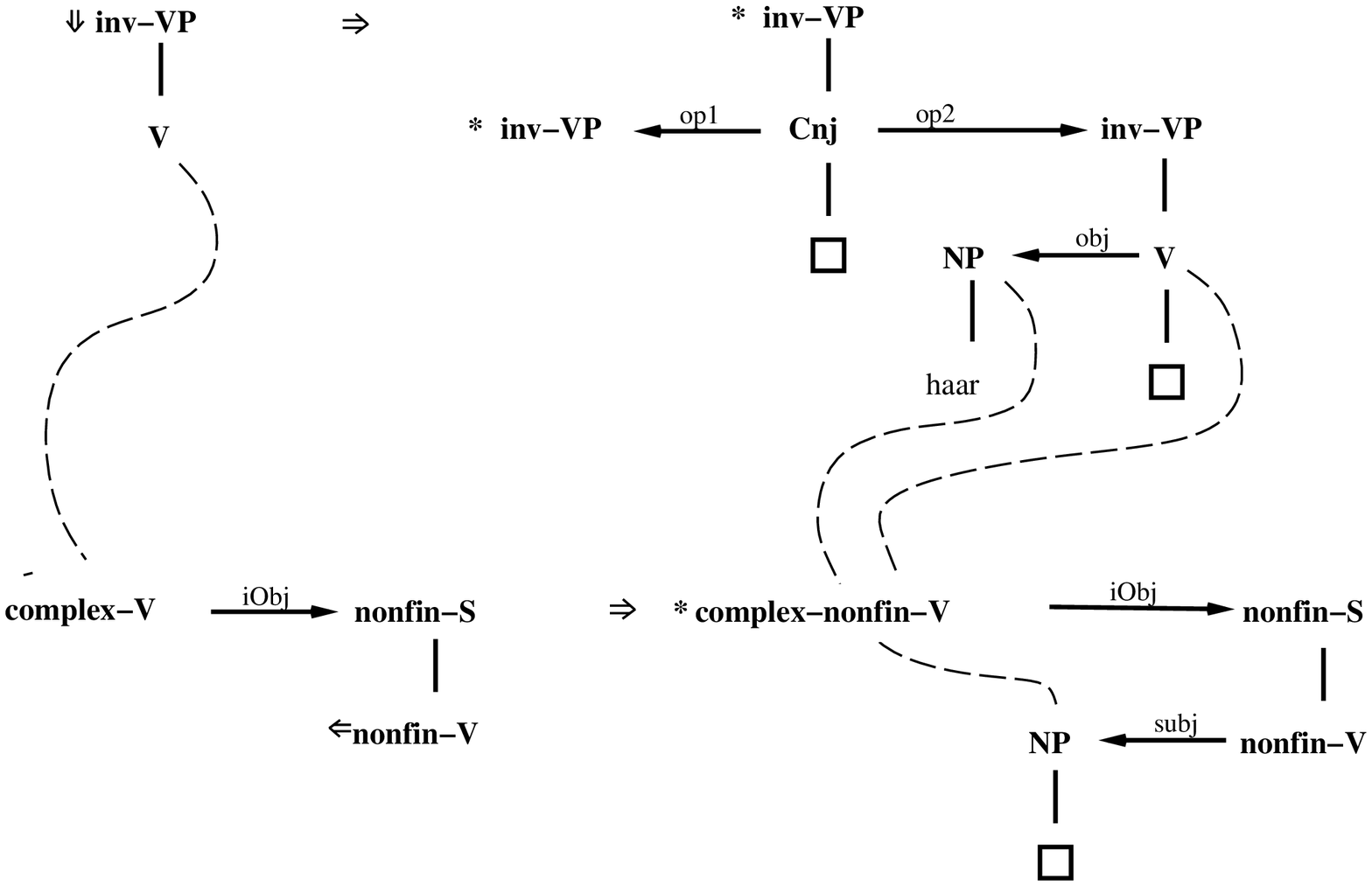,width=.9\textwidth} \end{center}
\caption{The rule for a nested verb of the {\em helpen} type}
\label{nestedComplexVP.fig}
\end{figure}

The rule represented on Figure~\ref{nestedComplexVP.fig} is a
double interpolation which acts simultaneously on both instances
of a split VP. A relation of coordination is supposed between the
inverted verb phrases on the left (top interpolation), for there is no clause
hierarchy among them. The use of double interpolation on two
graph areas connected by a coreference link allows these areas to
lie at an arbitrary distance from each other.

\begin{figure}[htbp]
\begin{center} ~\psfig{file=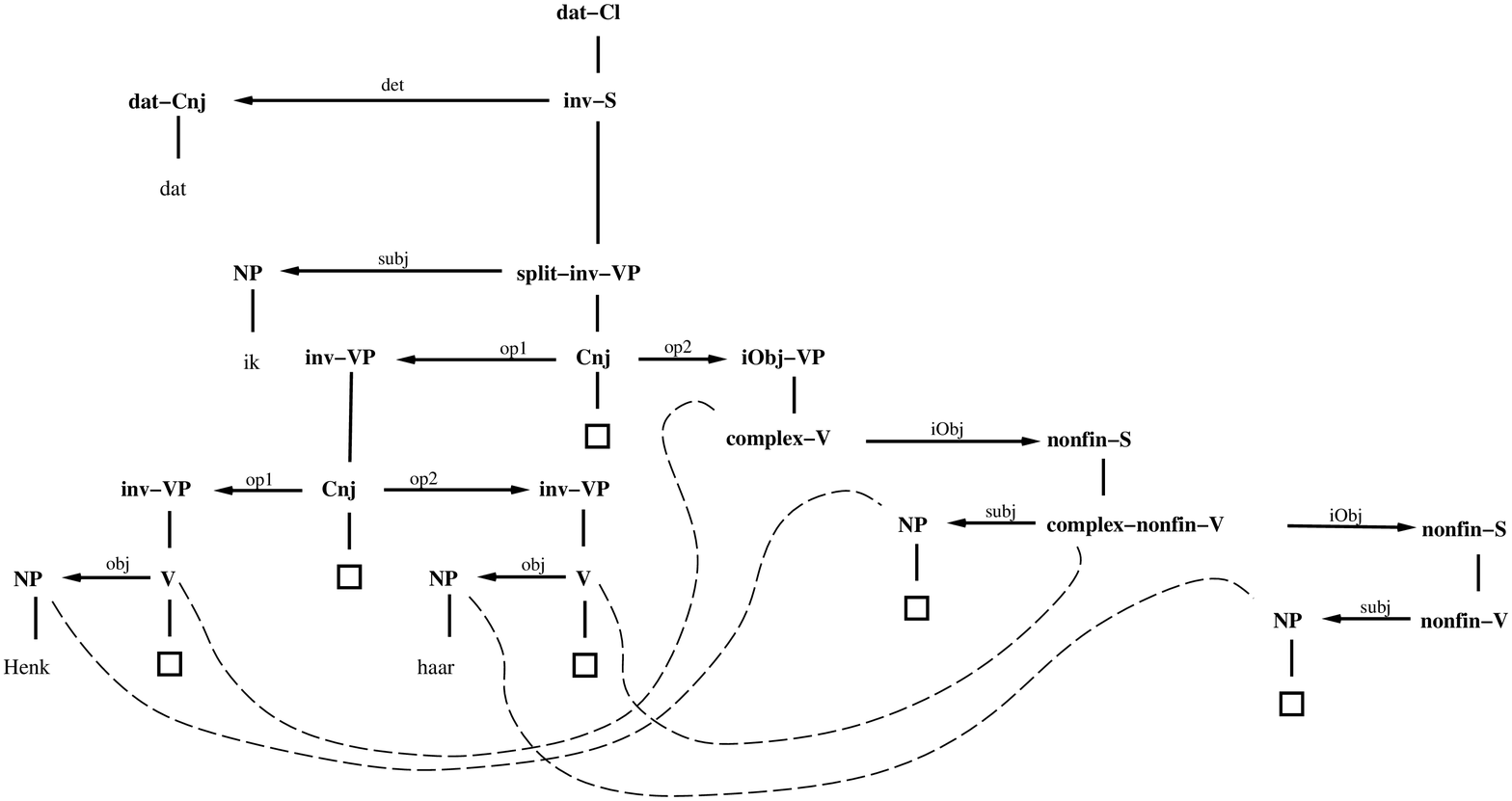,height=\textheight,angle=90} \end{center}
\caption{The output of the rule on Figure~\protect\ref{nestedComplexVP.fig}}
\label{graphBeforeNijlpaarden.fig}
\end{figure}

Supposing the lexemes {\em ik} and {\em Henk} were inserted into
the output of the rule represented on
Figure~\ref{complexDatClause.fig}, the effect of the last rule
would be to produce the graph on Figure~\ref{graphBeforeNijlpaarden.fig}.

\subsection{Simple nonfinite clause}

\begin{figure}[htbp]
\begin{center} ~\psfig{file=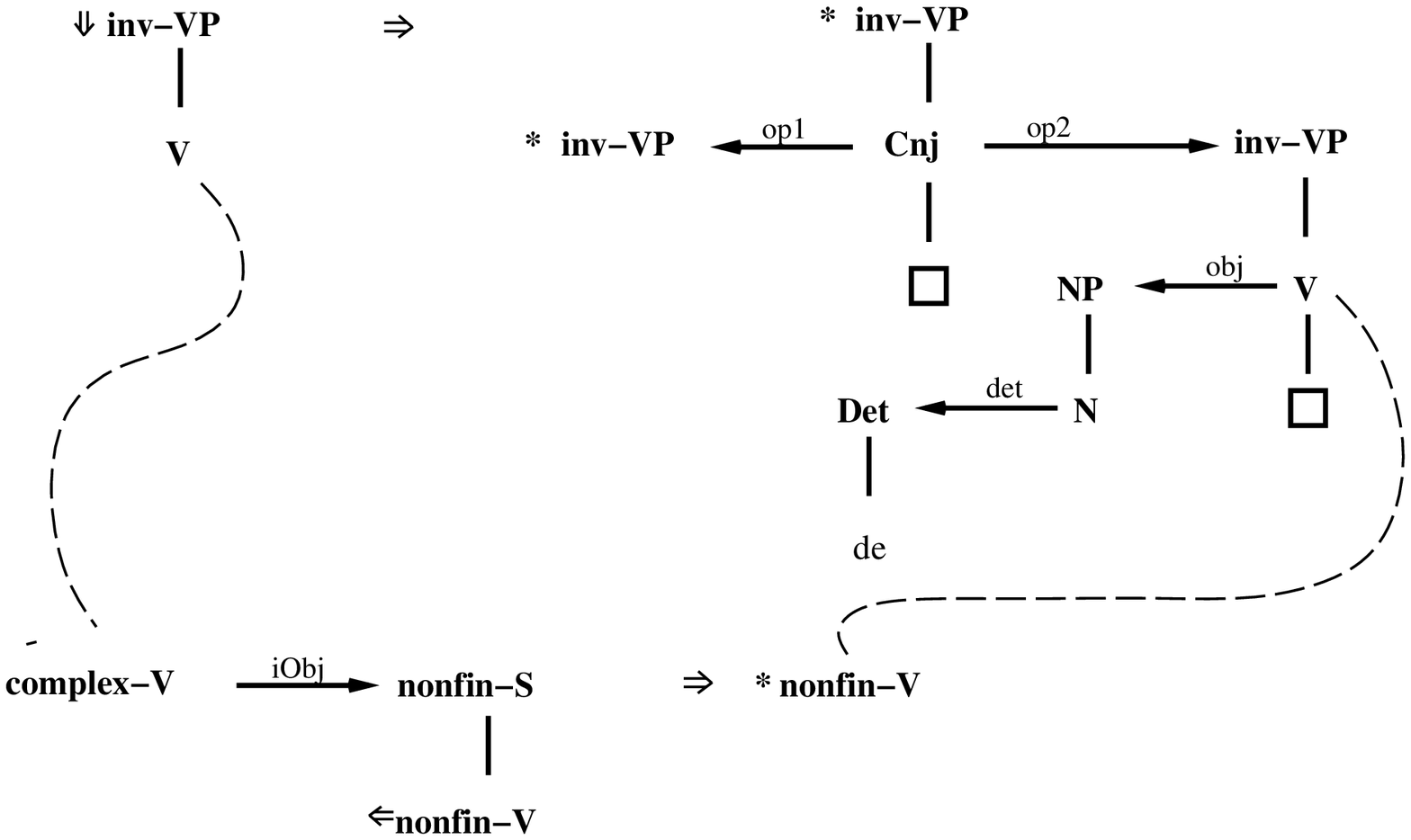,width=.8\textwidth} \end{center}
\caption{The rule for a nested monotransitive verb}
\label{nestedSimpleVP.fig}
\end{figure}

With a simple transitive verb, the verb gap that is created is
simply coreferenced to its counterpart in the context; so the
bottom interpolation does not create any node. The presence of an
article in the context specified on
Figure~\ref{nestedSimpleVP.fig} announces a transitive verb; and
so no rollbacking need be involved to trigger the rule that is
represented.

This rule adds an NP~node to the left of the ``syntactic rift'';
this node is the object of an {\em inv-VP} containing an anticipatory
verb gap coindexed with the second context anchor, which itself
anticipates the occurrence of the verb {\em voeren} to the right
of the rift. In short, the top interpolation adds a direct
object, and the bottom interpolation connects it to the cascade
of clausal objects after {\em zag}.

\section{Conclusion}

This \paper\ has defined a form of grammar, called \GIG, which
produces parse graphs by simple derivation from an
initial graph consisting of one node. Since derivation integrates
input lexemes left-to-right, a GIG derivation is an incremental
parse of an input sentence.

The basic operation, graph interpolation, which combines path
substitution and disjoint graph union, has been described at some
length. Two classic problems, one from the field of parsing
(arithmetic expressions), and one from the field of linguistics
(Dutch cross-serial dependencies), have been selected to
illustrate problem-solving through GIG~design.

The essential, hopefully, has been stated, but much remains to be
done to improve the formalism and explore its
potentialities. Areas for further research are presented in
Section~\ref{further.sec}.

On the other hand, the GIG model shares intuitions with several
other syntactic formalisms, which are briefly reviewed in
Section~\ref{related.sec}.

\subsection{Related works}
\label{related.sec}

Graph interpolation can be viewed as an extension of tree
adjunction to parse graphs. And, indeed, TAGs
\cite{joshi-etal-75}, by introducing a 2-dimensional formalism
into computational linguistics, have made a decisive step towards
designing a syntactic theory that is both computationally
tractable and linguistically realistic. In this respect, it is an
obligatory reference for any syntactic theory intent on
satisfying these criteria.

The basic intuition in GIGs, however, that of incrementally
connecting lexemes by looking up their combinatory properties,
can be found in link grammars~\cite{Sleator95}. Link grammars are
mathematically interesting, but do not easily interface with a
semantic component.

Categorial grammars \cite{Steedman86,Steedman88} have a similar
rationale. They use binary combinators on function types that
represent grammatical categories. The parse structures
generated, which are binary trees reflecting the order of
combinator application, can be somewhat counterintuitive from a
semantic point of view.

In Lexical Functional Grammars~\cite{Bresnan85}, grammatical
functions are loosely coupled with phrase structure, which seems
to be just the opposite of what is done in a GIG, in which
functional edges are part of the phrase structure. Nonetheless,
these two approaches share the concern of bringing out a
functional structure, even if much of what enters into an
f-structure (i.e. a functional structure) in LFG is to be
addressed by the semantic component ---a topic for further
research--- in GIG.

\subsection{Further research}
\label{further.sec}

\subsubsection{Linguistic phenomena}

Important categories of phenomena to analyze include ellipsis,
notably under coordination, and word scrambling.

Furthermore, the discussion of free word order in this \paper\
has hovered at an uncomfortable level of abstraction. It is
necessary to return to this issue with concrete examples.

On the other hand, the interface with a scanner and morphological
analyzer has yet to be stated explicitly. Phenomena to account
for include idioms, phrasal lexical items, agglutinative
morphology, and scanning ambiguities.

\subsubsection{Psycholinguistic considerations}

The main natural feature embodied in \GIGs\ is incremental
left-to-right parsing, which is expected to provide a basis for
incremental semantic evaluation of discourse, a feature in favour
of which there seems to be conclusive experimental evidence.

\subsubsubsection{Processing time}

The formalism would be particularly satisfactory as a model of
human parsing if the computational complexity it predicts
correlated with observed processing time in humans. What is
mainly at stake here is the plausibility of the backtracking
mechanism that has been described.

\subsubsubsection{Garden-path sentences}

GIG parsing involves backtracking for sentences that humans do not
identify as garden-path sentences. A hypothesis to test against
further evidence is that the human analyzer is garden-pathed only
when semantic interpretation has to be revised, not when purely
structural adjustments are taking place. A thorough discussion of
this question requires that a working model be proposed for the
semantic component.

\subsubsubsection{Error tolerance}

The capacity to pick up meaning from ill-formed utterances is a
capability of the human language processor that does not seem
beyond the reach of GIG-based modelling. The model could involve
some form of partial match of context patterns to resort to when
the parser is stymied.

\subsubsubsection{Evolutionary aspect}

Languages undergo idiolectal variations and historical changes. A
constant adjustment of rules takes place when language is
used. If GIGs were endowed with the error recovery device
delineated in the previous paragraph, a model of evolution could
then be worked out to promote heavily activated near-matches to
full matches, possibly at the detriment of competing weakly
activated full matches, which could disappear in the process.

\subsubsection{Mathematical properties}

\subsubsubsection{Expressive power}

GIGs are at least as expressive as TAGs, for graph interpolation
can be used to express tree adjunction or tree substitution. This
means that the formalism is at least mildly
context-sensitive. Furthermore, multiple interpolation adds
expressive power comparable to that found in
Multi-Component~TAGs~\cite{Kroch87a}. GIGs do seem to provide the
``right'' amount of context-sensitiveness, but this remains to be
precisely quantified.

\subsubsubsection{Complexity}

The backtracking mechanism induces a worst-case exponential
complexity. On realistic grammars, however, both the base of the
exponential ---i.e. the number of lexemes past which it makes
sense to backtrack--- and the exponent ---i.e. the number of
competing rules for a given lexeme--- seem to have very low upper
bounds. What is a ``realistic'' grammar could be characterized
precisely by determining such bounds by collecting linguistic
facts. Beside quantified bounds, it is to be thought that some
barrier effect makes it pointless to backtrack beyond certain
syntactic barriers. On the other hand, the backtracking mechanism
itself could be reconsidered to reduce its computational
complexity, for example by allowing backtracking to skip back to
a privileged place either through links built while parsing or
using a form of pattern matching.

\subsubsubsection{Automatic generation of GIG rules}

Section~\ref{cfg.sec} suggests that a \GIG\ could be
automatically derived from a Context Free Grammar, mostly for the
purpose of using existing LR~grammars.

The feasibility of converting automatically from other
formalisms, and particularly Lexical Functional Grammars, seems
worth exploring as well.

\subsubsubsection{Linearized syntax for rules}

From a practical point of view, however, GIGs are not to be seen
as a low-level formalism to be generated automatically from
higher-level formalisms. But, for GIG writing to be practical,
what could be needed is {\em (i)}~a linearized syntax to type
rules quickly rather than draw graphs, and {\em (ii)}~a WYSIWYG
tool that showed incrementally the graphic aspect of what is
being typed. (This tool could of course be implemented on top of
a GIG engine.)

A linearized syntax could involve a frame-like notation in which
an augmented parse graph were a set of slots with a {\em
phrases} slot that pointed (via slot values) to phrases with an
optional pre-defined {\em parent} slot and user-defined
functional slots.

For example, a linearized version of the addendum in
Figure~\ref{nounless.fig} could look like the frame expression on
Figure~\ref{LinearizedNounless.fig}. (Predefined slot names are
boldfaced. Slot references are slash-separated paths starting at
the parent of the slot being defined. A double dot (..) moves up
one level in the slot hierarchy.)

\begin{figure}[htbp]

\begin{tabbing}
xxxx\= xxxx\= xxxx\= xxxx\= xxxx\= xxxx\= \kill
{\bf addendum} \{ \+ \\ 
  {\bf phrases} \{ \+ \\
      0 \{ {\bf head} N; \}; \\
      1 \{  \+ \\
        {\bf parent} ../0/head; \\
        {\bf head} Adj; \\
        subj N; \- \\
      \}; \\
      2 \{  \+ \\
        {\bf parent} ../1/subj; \\
        {\bf head} Gap; \- \\
      \}; \- \\
  \}; \\
  {\bf lexeme} \{  \+ \\
    {\bf parent} ../phrases/1/head;  \\
    {\bf spelling} "wealthy"; \- \\
  \};          \\
  {\bf anchors} \{ \+ \\
    {\bf source} ../phrases/0; \\
    {\bf destination} source; \- \\
  \}; \\
  {\bf coreferences} \{ \+ \\
    0 \{ \+ \\
      0 ../../phrases/0/head; \\
      1 ../../phrases/1/subj; \- \\
    \}; \- \\
  \}; \- \\
\}
\end{tabbing} 
\caption{A linearized addendum}
\label{LinearizedNounless.fig}
\end{figure}

On the other hand, many grammar designers would probably not feel
frustrated by the lack of a linear syntax given an efficient
graphic tool.

\bibliography{strings,general}
\bibliographystyle{alpha} 

\appendix

\section{Anchor location procedures}
\label{locateAnchor.app}

The following procedure can be used to update the set of possible
linear successors of the current lexeme after an
interpolation. It is operational even when several linear orders
are allowed.

\begin{procedure} \mbox{} \\
\label{updateAnchorSet.proc}
\begin{enumerate}
\item Based on ordering relations in the addendum, one identifies
the set of possible linear successors of the current lexeme,
which is here the lexeme just inserted. 
\item The possible linear successors of the {\em previous}
lexeme, as recorded when it was inserted, minus the node that
matched the context anchor, are added to the set found in
step~1.
\end{enumerate}
\end{procedure}

Note that, since the addendum and the context are disjoint, the
second step cannot add potential successors already added by the
first step. This observation is important to determine whether rule
application remains reversible in the presence of free word
order. Indeed, undoing a rule with respect to the set of
potential successors of the current lexeme can be done as
follows.

\begin{procedure} \mbox{} \\
\begin{enumerate}
\item Based on ordering relations in the addendum, one identifies
a subset of possible linear successors to the current lexeme,
i.e. the lexeme on which to backtrack, and subtracts this subset from the
set of possible successors.
\item Once the insertion itself has been undone, the context
anchor for this insertion is added to the set obtained in the
previous step, and this set becomes the set of potential
successors of the previous lexeme. 
\end{enumerate} 
\end{procedure}

\end{document}